# Criticality Metrics for Automated Driving: A Review and Suitability Analysis of the State of the Art

Lukas Westhofen[1] · Christian Neurohr[1] · Tjark Koopmann[1] · Martin Butz[2] · Barbara Schütt[3] · Fabian Utesch[4] · Birte Neurohr[1] · Christian Gutenkunst[5] · Eckard Böde[1]

## Abstract
The large-scale deployment of automated vehicles on public roads has the potential to vastly change the transportation modalities of today's society. Although this pursuit has been initiated decades ago, there still exist open challenges in reliably ensuring that such vehicles operate safely in open contexts. While functional safety is a well-established concept, the question of measuring the behavioral safety of a vehicle remains subject to research. One way to both objectively and computationally analyze traffic conflicts is the development and utilization of so-called criticality metrics. Contemporary approaches have leveraged the potential of criticality metrics in various applications related to automated driving, e.g. for computationally assessing the dynamic risk or filtering large data sets to build scenario catalogs. As a prerequisite to systematically choose adequate criticality metrics for such applications, we extensively review the state of the art of criticality metrics, their properties, and their applications in the context of automated driving. Based on this review, we propose a suitability analysis as a methodical tool to be used by practitioners. Both the proposed method and the state of the art review can then be harnessed to select well-suited measurement tools that cover an application's requirements, as demonstrated by an exemplary execution of the analysis. Ultimately, efficient, valid, and reliable measurements of an automated vehicle's safety performance are a key requirement for demonstrating its trustworthiness.

The research leading to these results is funded by the German Federal Ministry for Economic Affairs and Climate Action within the projects 'VVM – Verification & Validation Methods for Automated Vehicles Level 4 and 5' and 'SET Level – Simulation-based Development and Testing of Automated Driving', based on a decision by the Parliament of the Federal Republic of Germany.

✉ Lukas Westhofen
lukas.westhofen@dlr.de

Christian Neurohr
christian.neurohr@dlr.de

Tjark Koopmann
tjark.koopmann@dlr.de

Martin Butz
martin.butz@de.bosch.com

Barbara Schütt
schuett@fzi.de

Fabian Utesch
fabian.utesch@dlr.de

Birte Neurohr
birte.neurohr@dlr.de

Christian Gutenkunst
christian.gutenkunst@avl.com

Eckard Böde
eckard.boede@dlr.de

[1] Institute of Systems Engineering for Future Mobility, German Aerospace Center (DLR) e.V., Oldenburg, Germany

[2] Robert Bosch GmbH, Renningen, Germany

[3] FZI Research Center for Information Technology, Berlin, Germany

[4] Institute of Transportation Systems, German Aerospace Center (DLR) e.V., Brunswick, Germany

[5] AVL Deutschland GmbH, Karlsruhe, Germany

# 1 Introduction

A launch of automated vehicles (AVs) to public roads promises various societal benefits, ranging from improving the economic efficiency of the transportation system up to increasing the mobility of parts of the population [23]. A key factor and potential bottleneck in this feat is assuring and demonstrating safe behavior of the driving functions deployed in such vehicles beyond functional safety [42]. Currently, publicly funded research undertakings, such as the PEGASUS family projects VVM and SET Level[1], are investigating and developing methods and tools for the verification and validation of functions implementing SAE Levels 4 and 5 [82].

Specifically, research interest lies within analyzing the emergence of *criticality* in traffic [69]. AVs are confronted with an unstructured, open context, particularly in mixed traffic environments. It is therefore imperative to identify potentially safety relevant factors of the context, such as the intricacies of human behavior, early on in the development process [85]. For automated driving functions, these factors can subsequently be drawn upon to implement safety mechanisms mitigating their effects, as the human driver is no longer available as a potential fall-back mechanism. Therefore, it is essential to possess the ability of computing the effects of such factors by means of appropriate measurement tools for criticality. We note that ambiguity exists as to how criticality is defined in both the scientific literature as well as in industrial practice [37]. For the scope of this work, criticality is understood as *'the combined risk of the involved actors when the traffic situation is continued'* [69].

As one piece to solving the previously sketched challenge, this paper examines *criticality metrics* for quantifying certain aspects of criticality. Similar to Schütt et al. [88], we represent a (scene level) criticality metric as a function $\kappa : \mathcal{S} \times \mathbb{R}^+ \to \mathcal{O}$ that measures, for a given traffic scene $S \in \mathcal{S}$ at a time $t \in \mathbb{R}^+$, aspects of criticality on a predetermined scale of measurement $\mathcal{O} \subseteq \mathbb{R} \cup \{-\infty, +\infty\}$. Scenario level criticality metrics extend this definition from scenes to scenarios [97], i.e. adding (retrospective) temporal aspects to the measurement. Most criticality metrics only quantify over a subset of the influencing factors that are associated with criticality, such as spatial, temporal, dynamical, perceptual, or environmental circumstances [29].

The capability of such metrics to compute an (aspectual) surrogate for criticality has lead to their potential being leveraged in various areas around automated driving, e.g. as objective functions for planning modules or for assessing the outcome of test cases. Due to the large amount of existing metrics as well as their variance in properties, selecting well-suited computational methods for criticality assessment within a given application poses an important challenge.

Hereon, this elaboration provides a blueprint to guide the reader in answering the question:

> How to identify a set of criticality metrics suited for computationally assessing criticality within the application at hand?

Therefore, we (a) *provide a novel method* guiding users in answering this question, and (b) *apply this method* based on a schematic, unifying review of the current state of the art of criticality metrics, their usage, and their features.

We emphasize the necessity of answering the aforementioned research question by the following illustration.

### 1.1 Motivating Example

Consider the well-known Time To Collision (TTC) metric, which intuitively queries a dynamic motion model (DMM) for a predicted collision between two actors and computes the time until said event. For example, the TTC can be used as a part of an automated emergency braking (AEB) system for car following scenarios, where the decision logic of the system can incorporate the metric's computation, e.g. by providing a warning to the driver [83]. Hence, within the application of an AEB, the TTC's information can play a vital role in situation assessment. By showing that its computed value provides a high sensitivity in car following scenarios w.r.t. the criticality induced by the lead vehicle, it can provide the AEB system with relevant information on the temporal proximity to a potentially necessary emergency braking. A highway AEB system that employs a TTC thus uses a reliable indicator in its decision making process.

As another example, the TTC can also suitably filter large highway data sets due to both its efficient algorithmic implementation as well as methods for aggregation over time that enable retrospective assessments. Such filtered data sets can then be used to build scenario catalogs for the verification and validation of highway driving functions. It has to be noted that it is unlikely that such a catalog contains every critical scenario, as the TTC only detects a certain aspect of criticality (cf. upper scenario in Fig. 1). For adequate coverage in identifying safety-critical highway scenarios in a data set, computing multiple metrics is therefore imperative.

Contrasting the previous applications, a TTC combined with a single point kinematics prediction model can assess criticality of intersection scenarios only partially, as it evaluates to infinity in many scenes. This problem arises if the DMM predicts no contact, which is likely for non-following scenarios: although it is probable that the *paths* of two vehicles intersect in such scenarios, it becomes unlikely for the actual *trajectories* to overlap, as proposed by Allen et al. in 1978 [4]. For this, the DMM would have to predict equal positions at the same time, which is, depending on the spatio-temporal representation of traffic participants, highly improbable.

This problem can be partially mitigated by employing suitable computational methods such as using point sets instead of single points within the DMM and a discretization of the geometrical space, effectively increasing the chance of trajectory intersection. As another viable approach, having

---

[1] https://vvm-projekt.de/en, https://setlevel.de/en

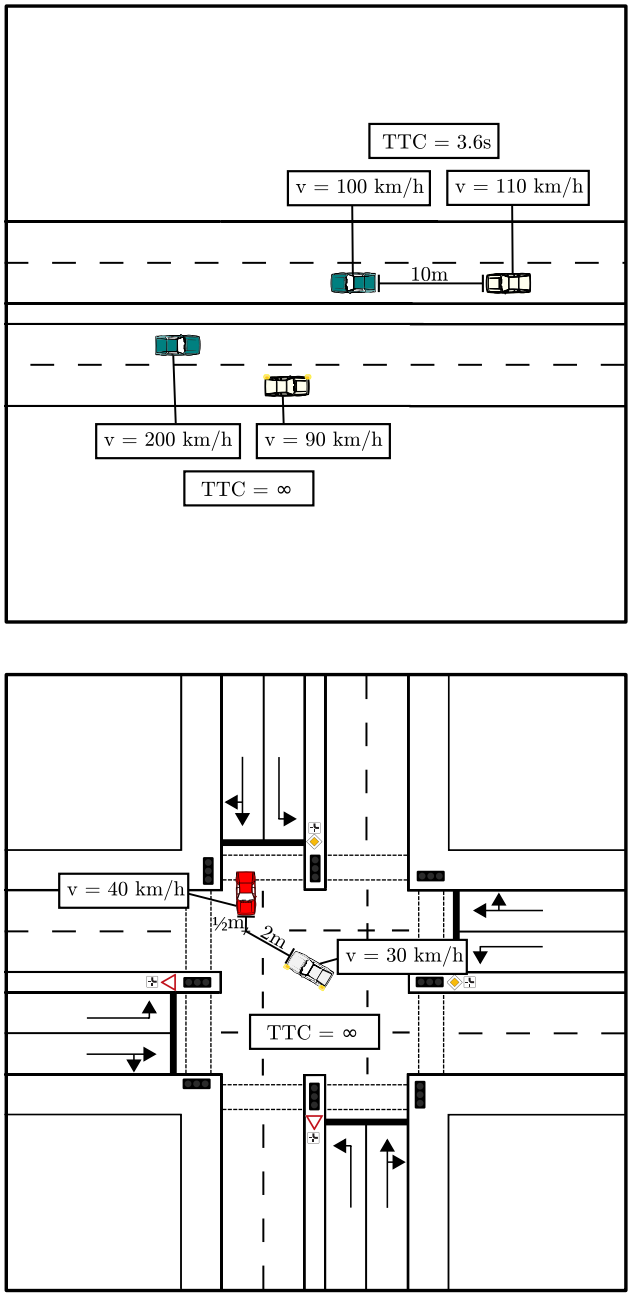

**Fig. 1** The TTC with a constant velocity model is a suitable metric for car-following scenarios (upper roadway in upper scene), but has greatly reduced sensitivity outside this scenario class, both on highways (lower roadway in upper scene) and intersections (lower scene)

models calculate multiple trajectories and searching for a collision [101] can increase the chance of predicting a contact. But, in its common form, the TTC with a single point kinematics suffers from reduced validity outside its designated scenario class, as depicted in Fig. 1. Thus, it is not advisable to solely rely on a TTC when computing a safety assessment of vehicles in urban intersections. For such applications, several metrics ought to be used in combination as to constitute an appropriate measurement.

### 1.2 Problem Statement

From this introductory example it becomes evident that, especially for safety-critical systems operating in open contexts, it is inevitable to perform an in-depth analysis of the match between the requirements imposed by the planned computation of criticality and the capabilities of the employed tools. This is specifically true when considering the use of traditional safety metrics originating from early traffic conflict research—such as the TTC—, where the primary design goal was the analysis of human behavior. Great care needs to be taken in selection and adaptation when utilizing such computational methods within the field of automated driving.

As to generalize the previously introduced challenge, one can identify the relevant sets of scenarios over the universe of all scenarios, $Sc$, given an application and a computation of a binary criticality classification:

1. the set of scenarios relevant for the application, $Sc_{app}$
2. the set of actually critical scenarios for the given application, $Sc_{app}^{crit}$
3. the set of actually uncritical scenarios for the given application, $Sc_{app}^{\neg crit}$
4. the set of critical scenarios as indicated by $\kappa$, $Sc_{\kappa}^{crit}$
5. the set of uncritical scenarios as indicated by $\kappa$, $Sc_{\kappa}^{\neg crit}$
6. the set of scenarios to which $\kappa$ is applicable, $Sc_{\kappa}$

The sets $Sc_{app}^{crit}$ and $Sc_{app}^{\neg crit}$ can be considered as the ground truth that is relevant for the application, which are in turn targeted to be measured by $\kappa$. An exemplary relation between those sets is depicted in Fig. 2.

Here, we are presented with an application that is concerned with a set of scenarios $Sc_{app}$, for example, the set of all highway scenarios. Let us assume that the metric requires high sensitivity, e.g. to computationally identify a large share of critical scenarios in a highway scenario data base. Formally, this means that the set $Sc_{app}^{crit} \setminus (\bigcup_{\kappa_i \in K} Sc_{\kappa_i}^{crit})$ shall be small. In general, besides the coverage of the scenario space, applications impose a variety of requirements on $K$, for example concerning the metrics' output scales or runtime capabilities.

The methodical approach and the state of the art review of this work provide a schematic solution to this generalized challenge of identifying a potent set of criticality metrics $K$ for $Sc_{app}$. This deliberate selection hence enables statements on the scenario coverage of criticality metrics.

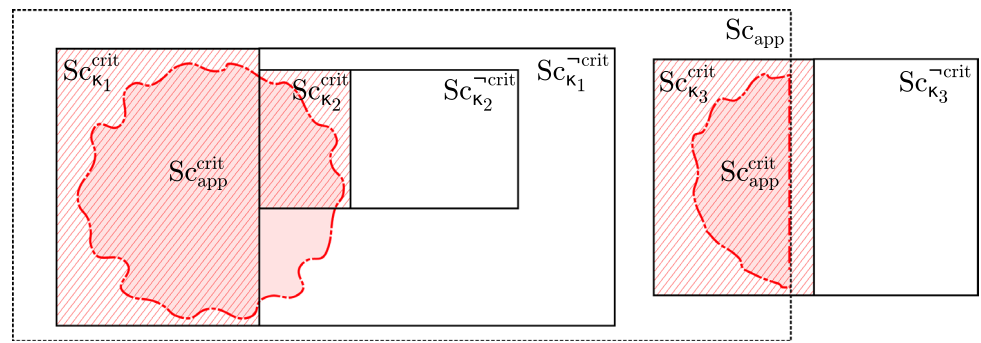

**Fig. 2** An exemplary, schematic scenario space $Sc_{app} \subseteq Sc$ for an application. The set of metrics $K = \{\kappa_1, \kappa_2, \kappa_3\}$ identifies certain scenarios inside Sc as critical and uncritical

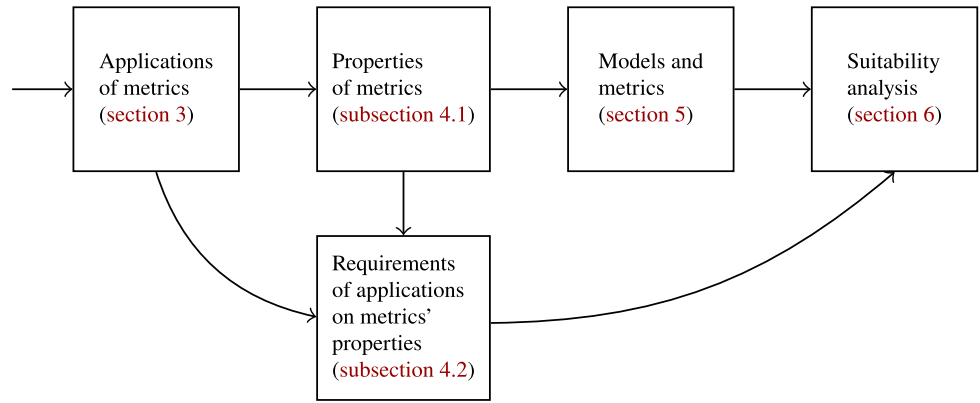

**Fig. 3** An overview of the method employed in this work

### 1.3 Overview of the Work

Systematic reviews of safety surrogates for traffic accident research have been conducted in the past and are referenced in Sect. 2. In contrast, this work focuses on their application to the domain of automated driving and both extensively reviews and unifies the state of the art. Specifically, the subject of the presented review and analysis is *single scenario level metrics*, assessing properties of interest considering the influence of the actors on the criticality of a single scene or scenario [88]. Here, traffic participants are considered a black box. There exist both metrics for more refined models as well as more abstract views on the traffic happenings. For the first, white and gray box approaches are proposed, e.g. to assess the performance of perception components, consolidated as *unit-level metrics*. In the second direction, there exist *accumulated scenarios level metrics* [88] to assess macroscopic properties [31]. Both approaches are not in the scope of this work.

Figure 3 sketches the methodical approach presented and utilized to review and analyze the state of the art within this publication.

We first present a review of contemporary applications of criticality metrics in Sect. 3. Note that our exemplary execution of the suitability analysis puts only rudimentary assumptions on the concrete shape of the applications, which affects the concretization of the results downstream. In practice, when confronted with a detailed application, the derived suitability claims will become more rigorous. In Sect. 4 we derive properties of interest that arise when considering the outlined applications. Subsequently, for each application, we identify requirements on the derived properties of the metrics. Taking this systematic requirement analysis into account, we review the state of the art of criticality metrics in Sect. 5. Here, we depict a large set of metrics and describe their basic concepts in a unified manner. Additionally, we examine to which degree the currently available metrics satisfy the previously derived properties. The results are available on a supplementary web page which is open for contributions from the community[2]. The suitability analysis that enables an evaluation of the review is subsequently depicted in Sect. 6. Here, we present an expert-based process to methodically identify potential metrics for a given application. Finally, the results of Sects. 4 and 5 are utilized to validate the proposed method, mapping an exemplary application's requirements on the metrics' actual capabilities.

To summarize, the *main contributions of this work* are to

1. give an extensive and unifying **review of the state of the art of criticality metrics**, their applications, and their properties,

---

[2] http://purl.org/criticality-metrics

2. provide a **blueprint for a suitability analysis** that uses the state of the art review as a backbone,
3. and **evaluate the review and proposed method** by means of examining an exemplary application.

Due to the rapid developments both in industry and research, there will prospectively exist new properties, applications, and metrics. The reader is therefore invited to conduct a custom suitability analysis based on the results presented and contribute gained knowledge to the open repository[2]. Consequentially, during the continuous pursuit of deploying safe automated vehicles on public roads, we imagine a constant extension and refinement of the catalog initiated in this work to readily identify a well-suited measurement tool for computing aspects of criticality within the task at hand.

## 2 Related Work

Safety surrogate indicators have been developed and employed to research and analyze traffic safety, e.g. for civil engineering and vehicle safety purposes. Firstly, the concept of analyzing *traffic conflicts* instead of solely relying on *traffic accidents* was popularized by General Motors in 1968 [76]. Canonically, the foundations to objectively identify and measure such conflicts have been laid in the 1970s, among others through the works of Hayward [33], Hydén [39], and Allen [4]. With a focus on traffic conflict analysis, several studies have reviewed the development of surrogate indicators since then [6, 65, 91, 99]. Besides reviewing safety surrogates, there also exist discussions on their relevant properties w.r.t. traffic accident research, such as reliability and validity [17, 59, 91], as well as requirements thereon [45, 91]. None of these studies is concerned with the application of such indicators to the domain of vehicle automation, which necessarily imposes a different perspective, e.g. regarding formal rigor or run-time capability.

In the AV domain, safety surrogate indicators are typically referred to as criticality metrics. An initial overview on contemporary criticality metrics has been published by the authors and is extended and unified in this work [69, p. 14]. Similarly, there exists work within the AV community that refers to a certain subset of criticality metrics, but does not provide an extensive review on the possible choices [31, 36]. Additionally, motion models play a crucial role when assessing future evaluations of a scene. In this regard, Lefèvre et al. have analyzed the suitability of different abstraction levels of motion models to be used for risk assessment [63], but focus on the modeling aspect and consider metrics only to a minor extent. Other studies present broader overviews on threat assessment methods for automated driving, but omit formal details and a concerned only with a restricted application setting [19, 90]. Additionally, the various advantages and disadvantages of the single approaches are not compared against each other, and no systematical approach for deriving a set of suitable metrics is given. The employment of such a method is essential when studying effects of combining various metrics given the goal of a reliable and valid measurement of criticality.

For the area of automated driving, this issue was partially recognized by Junietz [47], who systematically derived requirements on criticality metrics, albeit being primarily concerned with two applications, namely to

1. extrapolate a macroscopic risk statement that can be used in a safety argumentation of a given system, and
2. identify critical scenarios within data.

The derived requirements are later being used to develop a custom criticality metric satisfying the demands of both application. Scientific and industrial approach have recently started to use criticality metrics in various other applications besides the aforementioned ones, motivating the need for a re-consideration of the set of applications to derive requirements from. The work at hand therefore extends the idea to current approaches, specifically subsuming the scenario classification as well as the safety argumentation application as presented by Junietz.

## 3 Applications

In general, criticality metrics are employed to enable quantified statements over criticality, e.g. for mathematical modeling, computation, and analysis. Those statements can be used in various automated driving applications, specifically those relying on computational methods, such as simulation. Within this work, we understand an application as a specific process that is conducted during the design, implementation, analysis, or deployment of an AV.

This section presents a selection of common applications that rely on computation of criticality metrics. Major parts were derived based on earlier work of the authors [68], where an extensive literature review on processes within scenario-based testing approaches for automated vehicles was performed.

For most of the identified approaches, criticality metrics play a central role. Such roles were subsequently categorized and are depicted in this section. An overview of the identified applications, structured along the V-model [42], is given in Fig. 4. The depicted set of common applications eventually facilitates the derivation of substantiated properties of criticality metrics from the application perspective, depicted in Sect. 4.

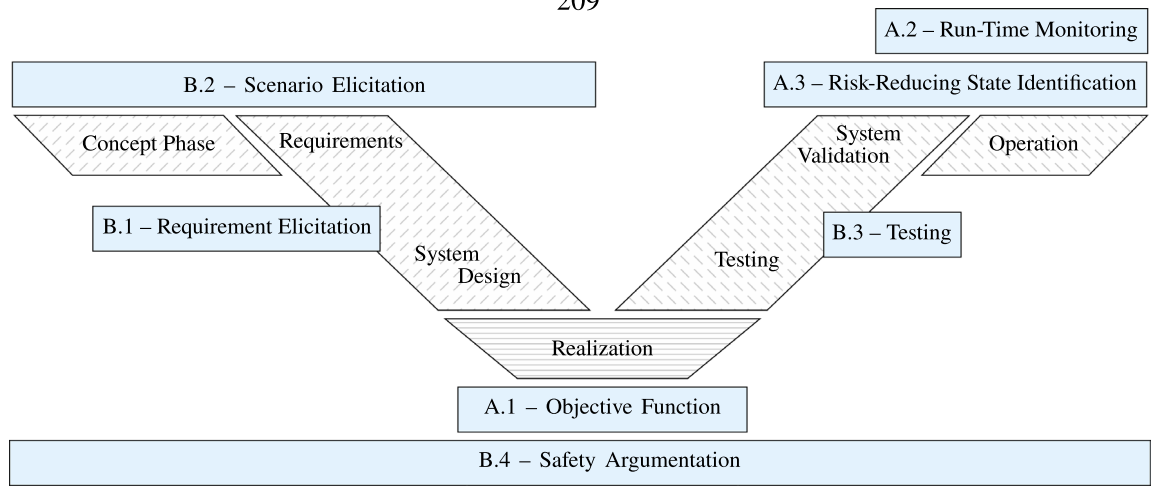

**Fig. 4** Arrangement of the identified applications A.1 to A.3 and B.1 to B.4 of criticality metrics (solid) along the V-model (hatched) [42]

Before beginning the literature review, we note that criticality metrics have been used under different terminologies in several fields of study and application areas over the last decades, ranging from traffic accident research over psychology up to vehicle automation, leading to a vast heterogeneity in the literature. Due to this, for assessing the state of the art, i.e. contribution 1), we deviated from the standard method of literature identification. Based on their diverse experience and collective knowledge in working in the aforementioned fields, the authors were able to assemble a list of applications, properties, and metrics. Due to this diversity in expertise, the identified state of the art is likely to be representative, i.e. to provide a sufficient coverage of the field. 9

## 3.1 Implementation

When implementing an automated driving function (ADF) to be used in an AV, a major aspect of the implementation's requirements is concerned with safety. One possibility of implementing safe driving functions is drawing on the computations of criticality metrics, as sketched in the following.

### 3.1.1 Objective Function (A.1)

An automated driving function can be formulated as a function optimization problem involving various encoded constraints regarding efficacy, comfort, and safety [28]. For the latter, directly or indirectly minimizing a criticality metric can be seen as one possibility to optimize safety. As an example, metrics have been used to purposefully increase the relevant training data set for planning components [1].

### 3.1.2 Run-Time Monitoring (A.2)

Monitoring the automated vehicle's state during run-time enables a dynamical assessment of the situational risk the vehicle is currently in [50], also known under the term *dynamic risk assessment* [80]. Based on these monitoring results, the system is able to then execute appropriate reactions, e.g. an evasive maneuver. For example, specifications can be formulated in a suitable logic, e.g. a signal temporal logic, which then accesses a criticality metric as a signal [104].

### 3.1.3 Identification of Risk-Reducing States (A.3)

In case a safety-critical state is detected, a fallback system can guide the vehicle to a state with lower risk, often called a minimal risk maneuver (MRM) [82]. In such a situation, other constraints—such as comfort—can be secondary. Here, the primary goal of the automated vehicle is be the identification of an optimally safe trajectory, e.g. by enforcing safety constraints to the planning optimization problem [93]. Similar to A.1, those constraints can be defined through appropriate criticality metrics.

## 3.2 Verification and Validation

In the following, we present applications that are used for verification and validation of AV safety. If only accidents are defined to be unsafe events, the set of examined scenarios is extremely small. In this case, a safety argumentation would be based on a small sample size, leaving room for large statistical variances. A criticality metric allows to increase the set of scenarios by also considering 'near-miss' events.

### 3.2.1 Requirement Elicitation (B.1)

*Defining Pass-/Fail-Criteria (B.1.a)* To evaluate test results, particular pass-/fail-criteria need to be derived from previously identified high-level safety goals [51], e.g. using a Systems Theoretic Process Analysis [26]. Together with target values [52], criticality metrics enable a precise argumentation on the performance of a vehicle in a test case.

### 3.2.2 Scenario Elicitation (B.2)

*Scenario Classification (B.2.a)* As to reduce downstream test efforts, the scenario space can be classified, e.g. in the form of a finite set of abstract or logical scenarios. This classification is performed relative to given criteria, e.g. the expected behavior of the ego vehicle [14] or the presence of phenomena [69]. Criticality metrics can be used complementary in the definition of scenario classes, e.g. to constrain a class to a certain measured aspect of criticality [54, 71].

*Scenario Instantiation (B.2.b)* Besides generating a set of meaningful abstract or logical scenarios, downstream testing requires representative instances of those classes. To identify particularly safety-relevant test cases, the choice of representation can be guided by criticality metrics [49]. One possibility of identification is the use of optimization methods to derive relevant parameter combinations [31]. The solution to this problem can be indirectly approximated by optimizing the scenario class through learned regression models of a criticality metric [69] or directly, e.g. through importance sampling or Bayesian optimization [26] w.r.t. to a criticality metric [9].

*Data-Driven Scenario Elicitation (B.2.c)* Scenario classes are often derived from data bases. Naturally, criticality metrics can be computed on such data.

(i) *Selective Data Recording (B.2.c.i)*: Conducting real world drives of test vehicles is one method to collect relevant data for verification and validation [106]. As critical situations during such drives are rare, identifying them through appropriate metrics directly during the time of recording can reduce the amount of stored data significantly [101].

(ii) *Data Filtering (B.2.c.ii)*: Once data has been collected, e.g. through naturalistic driving studies or real world drives, one can identify particularly safety-relevant data samples, e.g. through querying and filtering the data basis [79]. As manual investigation is cumbersome, criticality metrics present as a tool in an automated analysis of large data sets [55].

Additionally, filtered data may also be clustered and compared to each other [84]. Those samples can later be used to build a scenario catalog [53].

### 3.2.3 Testing (B.3)

*Search-Based Testing (B.3.a)* When executing an abstract test case, it is of interest to guide the system under tests into states that challenge it w.r.t. to the tested property [32]. If we are given a safety property, criticality metrics can be employed as a continuous or even differentiable fitness function to effectively test the most safety-relevant aspects [13]. This can be achieved through (bounded) search-based optimization approaches [27].

*Test Evaluation (B.3.b)* During or after the execution of a test case, the performance of the system under test needs to be evaluated. This can be done based on the previously defined pass-/fail-criteria, which involve—possibly multiple—criticality metrics indicating the AV's safety performance [36]. In contrast to defining pass-/fail-criteria, a test evaluation actually computes the value of metrics and compares their results against the given criteria. Additionally, tests may not be evaluated dichotomously, but also rated according to more complex scales.

### 3.2.4 Safety Argumentation (B.4)

*Quantification for Hazardous Situations (B.4.a)* Metrics can be used in a safety case to support a quantified argumentation on the automation's safety. For example, they can be utilized to demonstrate the effectiveness of a risk mitigation strategy for a given hazard. Specifically, this can be achieved by showing to which degree the strategy reduces the criticality of such hazardous situations [52].

## 4 Properties of Criticality Metrics

This section firstly presents properties of criticality metrics that are derived from the identified applications of Sect. 3. Subsequently, we identify requirements that the applications impose on those properties. In the end, the requirements on the properties become relevant when choosing a criticality metric for a given application. Thus, they serve as the basis for the suitability analysis in Sect. 6. Note that there obviously exists a large amount of inter-dependencies between the properties, which we explicitly omit for sake of readability.

### 4.1 Derivation of Properties of the Metrics

For each property, we state a short explanation and an illustration. Additionally, if possible, we give an exemplary scale on which the metrics can be assessed on, which is then used in the subsequent requirement mapping. Note that the scale is employed for the purpose of this paper and may be tuned in granularity as seen fit.

### 4.1.1 Run-Time Capabilities

Some of the previously presented applications are real-time based. Hence, they require metrics to be computed during the application's run-time, i.e. online. This implies that the metric's algorithm is receiving an input data stream. This property subsumes a variety of run-time related sub-properties, for example those that are concerned with sampling rates or deadline responsiveness. In our case study, the property is measured on a binary scale. As an example, live data filtering in a measurement vehicle requires run-time capabilities. Here, if no buffer is implemented, the metric computation needs to be finished before the next data sample comes in.

### 4.1.2 Target Values

Some applications require the metric results to be compared to target values[3]. Such target values are highly dependent on the subject type, where for example a TTC limit can vary based on whether a human-driven or automated vehicle is assessed. This property states that there exists research, suggestions, or standards on potential target values for the given application and subject type. We assess the property on a binary scale for the scope of this publication. To give an illustration, the definition of pass-/fail-criteria requires to research target values to confidently assess the test performance.

### 4.1.3 Subject Type

Safety indicators have been developed with various subjects in mind, e.g. specifically for pedestrians. Therefore, we are concerned with the types of subjects criticality metrics will be used on by the application. A subject is defined as the entity for which the criticality metric is computed for. We differentiate between metrics being applied to human subjects and automations on an abstract scale [101]. This distinction becomes especially important when considering the environment of the AV: for mixed traffic applications, one needs metrics that assess the criticality of human behavior as well that of automated vehicles. In fully automated traffic, this is no longer necessary. Then, metrics that are applicable solely to AVs can be used to estimate safety w.r.t. to the other actors. To give an example, analyzing a driving study of human drivers to mine relevant scenario for a safety assurance in mixed-traffic requires metrics specifically suitable to assess human behavior.

### 4.1.4 Scenario Type

The types of scenarios for which criticality metrics are computed differ widely from application to application. For example, an AEB system employs criticality metrics on highway following-vehicle scenario types. On the other hand, the analysis of an urban intersection data set widens the scenario type. This property defines for which types of scenarios the application wants to use the metric. Scenario types can vary depending on the application, we therefore omit an assessment scale at this point.

### 4.1.5 Inputs

In the open contexts that AVs operate in, there exists a large set of criticality-relevant entity types, such as various types of road infrastructure, traffic rules, and dynamic objects. Criticality metrics reflect certain aspects of those entities. Naturally, the question arises which entities are necessary to give as inputs for the computation of a criticality metric. This may concern rudimentary information such as positions and speed for simple trajectory data sets, but can increase in complexity up to road geometry, types and amount of dynamic objects (cars, pedestrians, etc.), and weather. Depending on the application at hand, there will only be certain inputs for the metrics available, e.g. a run-time monitor has only access to information provided by the AV's perception subsystem. We define a possible input of a criticality metric to be a subset of all traffic entities and their properties, e.g. all dynamic objects $o$ with their velocity $o.v$. Furthermore, we note that metrics can impose restrictions on its own inputs, e.g. $o.v \neq 0$.

### 4.1.6 Output Scale

Applications will obviously use the outputs of the metric's computation. Depending on the subsequent utilization of the result, certain restrictions can be imposed on the output scale. This especially concerns the algebraic operators applied on the result. The output scale can be defined through the use of algebraic operators into nominal, ordinal, interval, or ratio scales, alongside potential physical quantities. To illustrate the concept, an optimization algorithm requires at least a partial order relation on the output. It can hence use no metrics whose results are given on a nominal scale.

---

[3] As in UL 4600 [96], i.e. a desired value, limit or incident frequency.

### 4.1.7 Reliability

The reliability of a measurement is defined as the degree of closeness of repeated measurements to one another and relates to the stability of the conclusions that can be drawn from the measurements [34, 57]. More specifically, highly reliable measures have a low random error and produce measurements that are consistent with each other. As criticality metrics are often defined through deterministic algorithms that are used in deterministic simulation environments, repeating their computations on the same inputs always yields the same outputs. Hence, our case is concerned with a certain facet of reliability, namely, in the difference of $\kappa(S)$ to $\kappa(S')$ for a scene $S'$ that has only a slight change in criticality compared to $S$. For an application, it can be crucial to be aware of the reliability of the employed metrics, e.g. when computing criticality over time in a simulation run. As another example, a safety case requires a metrics with high reliability—in the general sense—due to the need for external reproducibility, e.g. by regulatory authorities.

### 4.1.8 Validity

Besides the repetition of measurements, various applications are also directly concerned with the validity of the measurement itself [34]. Here, validity is defined as the closeness of the metrics' measurement to representing the actual accident probability and severity [57]. The validity of a metric can often only be stated w.r.t. the metric's scenario type—typically, using a metric outside its designated scenario type is expected to result in decreased validity. Note that a highly valid metric is inherently reliable, as a high validity directly implies the impact of a random error being low. A highly reliable metric on the other hand is not necessarily always a valid metric. As to give an example, a planning function necessitates highly valid metrics to be used in safety-relevant decision making. The validity of binary classifiers can for example be assessed on a [0, 1] scale using an accuracy score. Non-binary metrics can be validated by means of statistical distance metrics to compare the distributions. At an extreme, if a metric is not defined for an input $x$, then we define it as invalid for $x$.

### 4.1.9 Sensitivity

In combination with given target values, a metric can induce a confusion matrix when evaluated against a ground truth [31]. The following two properties refer to such a matrix. Note that a valid ground truth is necessary for an assessment. Sensitivity, also called the true positive rate (TPR), is defined as the rate of correctly identified critical situations, i.e. the share of true positives (TP) among the sum of TP and false negatives (FN):

$$TPR = \frac{TP}{TP + FN}.$$

An over-approximating criticality metric will induce a sensitivity value of one, as it correctly captures all of the critical situations (but obviously identifies many uncritical ones as critical). In an application such as data filtering one may require such over-approximating metrics in a first step [101]. Sensitivity is measured as a fraction, i.e. a value in [0, 1].

### 4.1.10 Specificity

Analogously to sensitivity, specificity, or the true negative rate (TNR), is defined as the rate of correctly identified uncritical situations, i.e. the share of true negatives (TN) among the sum of TNs and false positives (FP):

$$TNR = \frac{TN}{TN + FP}.$$

When using criticality metrics, one is often not only concerned with identifying critical situations but also having guarantees on a given scenario being uncritical. As an illustration, a planning component of an AV can require certain guarantees on the metrics' specificity. On the other hand, applications such as data filtering do not impose severe restrictions on specificity, although the filtering process efficacy improves with increased specificity. Analogously to sensitivity, specificity is measured as a fraction in [0, 1].

### 4.1.11 Prediction Model

Most scene level metrics rely heavily on models that allow to assess the criticality based on the predictions of the future states of individuals within the current scene. Note that it is also possible for scenario level metrics to apply a prediction model for each time point retrospectively, e.g. when aggregating the TTC over time. The prediction model property can be characterized in two dimensions:

1. the size of the time window allowing useful predictions (unbound or $x$ seconds), and
2. whether the metric only considers a single, possibly the worst case, evolution (linear time) or multiple parallel lines of development (branching time).

For example, an identification of risk-reducing states requires that the valid prediction horizon should be bound to at least the predicted duration of the planned risk-reducing maneuver. Additionally, the safety of the maneuver is increased when multiple future developments are considered, e.g. by incorporating possible reactive behaviors of other traffic participants to the maneuver. A prediction model is typically bound to a specific entity type, e.g. being only applicable to predict the behavior of a human passenger car driver. When selecting a prediction model, the type of the model has to fit the entity under consideration. Illustrating this, the behavior of an automated vehicle may not be validly predicted based on human factors such as human reaction times.

### 4.2 Mapping Requirements of Applications on Properties of Metrics

Based on the previously identified applications and relevant properties of criticality metrics, we derive requirements from the first on the latter. For conciseness, this mapping is depicted in Table 4 in B. Here, expert-based hypotheses formulated by the authors are presented. Due to the hypothetical nature, they are obviously subject to scientific debate, especially when presented with a refined application. Nevertheless, we show that even those abstract applications impose certain qualitative requirements.

## 5 Models and Metrics

There exists a wide arsenal of functions measuring various aspects of traffic safety.

Collecting their state of the art is the final preparatory step for executing the suitability analysis method presented in this work and instantiated in Sect. 6.

This section therefore presents a large set of metrics and identifies the features of their properties. Due to the exemplary nature of this process and the little assumptions on the implementation of the metrics, we are confronted with a high level of abstraction. Thus, it is impossible to instantiate all metric properties on a concrete level. This holds specifically for reliability, validity, sensitivity, and specificity, as they are dependent on specific implementation details of the metric models and algorithms. In fact, this leads to properties being seemingly intangible at this high level of abstraction. We reiterate that these are nonetheless paramount for both the presentation of the methodical approach as well as a consideration in a concrete, practical instantiation of the presented suitability analysis. In the latter case, the properties will be instantiated with more tangible features.

For simplified readability, common abbreviations and variables used in the definitions are aggregated in Table 1. Additionally, the index of acronyms of criticality metrics can be found in A.

### 5.1 Employed Models

Before the examined criticality metrics are described in detail, we present some shared, common models of the presented metrics. Such predictive models are applied to generate possible future developments of a scene. Note that the validity of the employed model can heavily influence a metric's measurements by invalidly representing properties of the entity under consideration, e.g. using human reaction times to model automated vehicle (AV) behavior or allowing an entity to directly achieve negative speeds after braking maneuvers.

When applying such predictions to the real world, one will have to deal with different uncertainties, e.g. process or measurement noise. These precision issues are sometimes modeled by adding white noise to the measurements or applying more sophisticated Kalman filter methods to the motion model. An additional factor that can reduce accuracy of motion predictions stems from the required discretization of the predicted motion. For more in depth considerations, one can also distinguish between different abstraction levels in motion modeling, specifically *interaction-aware*, *maneuver-based*, and *physics-based* models [63]. However, for the scope of this paper, we will not follow this refined contemplation and not consider models without noise, uncertainties, or discretization errors. Additionally, when calculating metrics from the local perspective of an actor, one can only measure relative motion. For the sake of brevity, the work at hand focuses on motions in a fixed global coordinate system. We refer to Schubert et al. for an example of using local coordinate systems [87].

*Single point kinematics* In a single point kinematics model, the development of motion variables over time is approximated by a Taylor polynomial [43]. Since this model is defined per actor, indices related to the actor are omitted. For position, velocity, and acceleration of an actor, it is defined as

$$\boldsymbol{p}(t + t_0) \approx \boldsymbol{p}(t_0) + \boldsymbol{v}(t_0)t + \frac{1}{2}\boldsymbol{a}(t_0)t^2 + \cdots + \frac{1}{n!}\boldsymbol{p}^{(n)}(t_0)t^n$$
$$\boldsymbol{v}(t + t_0) \approx \boldsymbol{v}(t_0) + \boldsymbol{a}(t_0)t + \cdots + \frac{1}{(n-1)!}\boldsymbol{v}^{(n-1)}(t_0)t^{n-1}$$
$$\boldsymbol{a}(t + t_0) \approx \boldsymbol{a}(t_0) + \boldsymbol{j}(t_0)t + \cdots + \frac{1}{(n-2)!}\boldsymbol{a}^{(n-2)}(t_0)t^{n-2},$$

where the power with brackets $(n)$ denotes the $n$-th derivative.

*Simple car model* The simple car model takes the speed and steering angle functions input vector $\boldsymbol{u} = (u^s, u^\phi) : \mathbb{R} \to \mathbb{R}^2, t \mapsto (u^s(t), u^\phi(t))$, and outputs the position and direction of the vehicle [61]. It takes the form

$$\dot{x} = u^s \cos(\psi), \dot{y} = u^s \sin(\psi), \dot{\psi} = \omega = \frac{u^s}{L} \tan(u^\phi),$$

where $L$ denotes the distance between front and rear axle. It is possible to add further derivatives by e.g. replacing $u^s$ with another unknown function whose derivative is given by a new control input $u^a$ specifying the acceleration [61]. This approach leads to an increased required regularity of the solution, while also increasing the ease of specifying more differentiable solutions if one keeps the differentiability of the control input functions the same.

**Table 1** Abbreviations and variables used for the presented metrics

| Identifier | Meaning |
|---|---|
| S | A scene |
| Sc | A a scenario |
| *Contextual information* | |
| $\mathcal{A}$ | Set of all actors in a scene or scenario |
| $A_i$ | Actor $i$ |
| $t_0$ | Starting time of a scenario |
| $t_e$ | Ending time of a scenario |
| $t$ | A point in time |
| $t_H$ | A time horizon |
| CA | A conflict area |
| *Entity- and time-dependent information* | |
| $\boldsymbol{p_O}(t)$ | Position of object $O$ at time $t$ |
| $\boldsymbol{p_i}(t)$ | Position of actor $i$ at time $t$ |
| $\boldsymbol{p_{i,m}}(t, t')$ | Position of actor $i$ at time $t$ when conducting maneuver $m$ at time $t$ |
| $d(\boldsymbol{p_1}(t), \boldsymbol{p_2}(t))$ | Euclidean distance of $\boldsymbol{p_1}(t)$ and $\boldsymbol{p_2}(t)$ |
| $\dot{d}(\boldsymbol{p_1}(t), \boldsymbol{p_2}(t))$ | Derivative of euclidean distance $d$ |
| $\boldsymbol{v_i}(t)$ | Velocity of actor $i$ at time $t$ |
| $\boldsymbol{a_i}(t)$ | Acceleration of actor $i$ at time $t$ |
| $\boldsymbol{a_{i,min}}(t)$ | Minimal available acceleration of actor $i$ at time $t$ |
| $\boldsymbol{a_{i,max}}(t)$ | Maximal available acceleration of actor $i$ at time $t$ |
| $\boldsymbol{j_i}(t)$ | Jerk of actor $i$ at time $t$ |
| $\boldsymbol{u_i}(t)$ | Control inputs of actor $i$ at time $t$ |
| $\boldsymbol{\beta_i}(t)$ | Sideslip angle of actor $i$ at time $t$ |
| $\boldsymbol{\phi_i}(t)$ | Steering angle of actor $i$ at time $t$ |
| $\boldsymbol{\theta_i}(t)$ | Steering rate of actor $i$ at time $t$ |
| $\boldsymbol{\psi_i}(t)$ | Yaw angle of actor $i$ at time $t$ |
| $\boldsymbol{\omega_i}(t)$ | Yaw rate of actor $i$ at time $t$ |
| $F_{idxy}$ | Tire forces of actor $i$ with direction $d$ for tire $(x, y)$ |
| $c_{i\alpha f}$ | Front tire cornering stiffness of actor $i$ |
| $c_{i\alpha r}$ | Rear tire cornering stiffness of actor $i$ |
| $l_{if}$ | Distance from front axle to center of gravity of actor $i$ |
| $l_{ir}$ | Distance from rear axle to center of gravity of actor $i$ |
| $L$ | Distance from front to rear axle |
| $m_i$ | Mass of actor $i$ |
| $I_{iz}$ | Moment of inertia of actor $i$ |
| $\delta_{if}$ | Front steering angle at the tires of actor $i$ |
| *Mathematical symbols* | |
| $\tau$ | A target value |
| $\|\cdot\|_2$ | The euclidean norm |
| $v_{long}$ | Longitudinal component of a vector $\boldsymbol{v}$ |
| $v_{lat}$ | Lateral component of a vector $\boldsymbol{v}$ |

*Continuous steering car* As an example to the previously mentioned idea, LaValle introduces the continuous steering car [61]. While solutions to the simple car state equation might have discontinuities in the steering behavior, by introducing the steering angle $\phi$ as a state variable and the steering rate $u^\theta$ as an input variable, we obtain

$$\dot{x} = u^s \cos(\psi), \dot{y} = u^s \sin(\psi), \dot{\psi} = \frac{u^s}{L} \tan(\phi), \dot{\phi} = u^\theta,$$

where the steering rate input $u^\theta$ might include discontinuities, but the steering angle $\phi$ does not. Notice that in this case, the input can be understood as $\boldsymbol{u} = (u^s, u^\theta) : \mathbb{R} \to \mathbb{R}^2, t \mapsto (u^s(t), u^\theta(t))$. Naturally it is also possible to apply this approach multiple times to further increase the required regularity of solutions.

*Coordinated turn model* The transition of the coordinated turn model without uncertainties [43] from a time step $t$ to a time step $t + T$ is given by

$$\begin{pmatrix} p_x^{t+T} \\ p_y^{t+T} \\ v_x^{t+T} \\ v_y^{t+T} \end{pmatrix} = \begin{pmatrix} 1 & 0 & \frac{\sin(\omega T)}{\omega} & \frac{1-\cos(\omega T)}{\omega} \\ 0 & 1 & \frac{\cos(\omega T)-1}{\omega} & -\frac{\sin(\omega T)}{\omega} \\ 0 & 0 & \cos(\omega T) & \sin(\omega T) \\ 0 & 0 & -\cos(\omega T) & \sin(\omega T) \end{pmatrix} \begin{pmatrix} p_x^t \\ p_y^t \\ v_x^t \\ v_y^t \end{pmatrix}.$$

Since the nearly coordinated turn model only differs by a uncertainty on the turn rate $\omega$, it is omitted at this point.

*Augmented coordinated turn model with polar velocity* In comparison to the last model, the augmented coordinated turn model also takes the turn rate as a state variable into account [81]. Furthermore, it demonstrates the possibility to use polar coordinates for velocities in prediction models. The underlying differential equation reads as

$$\begin{pmatrix} \dot{p}_x \\ \dot{p}_y \\ \dot{v}_{long} \\ \dot{\phi} \\ \dot{\omega} \end{pmatrix} = \begin{pmatrix} v_{long} \cos(\phi) \\ v_{long} \sin(\phi) \\ a_{long} \\ \omega \\ \alpha \end{pmatrix},$$

where $\alpha$ denotes the rotational acceleration. For constant velocity and zero rotational acceleration, we obtain the augmented coordinated turn model as a solution of the differential equation as

$$f\left(\begin{pmatrix} p_x \\ p_y \\ v_{long} \\ \phi \\ \omega \end{pmatrix}\right) = \begin{pmatrix} p_x + \frac{2v_{long}}{\omega} \sin(\frac{\omega T}{2}) \cos(\phi + \frac{\omega T}{2}) \\ p_y + \frac{2v_{long}}{\omega} \sin(\frac{\omega T}{2}) \sin(\phi + \frac{\omega T}{2}) \\ v_{long} \\ \phi + \omega T \\ \omega \end{pmatrix}.$$

For non-zero linear or rotational acceleration, solutions to the differential equation will obviously become more complex.

*One track model* The one track model assumes the speed of the center of gravity of the vehicle along the planned trajectory to be constant and the tires to be jointly modeled as a single tire at the center of both axles [86]. Since this model is defined per actor, indices related to the actor are omitted. The state update equation for this model can be formulated as

$$\begin{pmatrix} \dot{\beta} \\ \dot{\omega} \end{pmatrix} = \begin{pmatrix} -\frac{c_{\alpha f}+c_{\alpha r}}{mv} & \frac{c_{\alpha r}l_r - c_{\alpha f}l_f}{mv^2} - 1 \\ \frac{c_{\alpha r}l_r - c_{\alpha f}l_f}{I_z} & -\frac{c_{\alpha f}l_f^2 + c_{\alpha r}l_r^2}{I_z v} \end{pmatrix} \begin{pmatrix} \beta \\ \omega \end{pmatrix} + \begin{pmatrix} \frac{c_{\alpha f}}{mv} \\ \frac{c_{\alpha f}l_f}{I_z} \end{pmatrix} \delta_f.$$

*Two track model* The two track model extends the one track model by dynamics for each individual tire and lifts the constant speed assumption [86]. Again, indices related to the actor are omitted. The state update equation reads as

$$\dot{\boldsymbol{x}} = f(\boldsymbol{x}, t, \boldsymbol{u})$$

where $\boldsymbol{x}$ is a 36-dimensional state vector including for example the position, cardan angles and vertical movements of the tires in a local coordinate frame and $\boldsymbol{u}$ is a 4-dimensional input vector with the steering angle, gas and braking pedal depression percentage, and a gear.

*Potential-based models* A potential-based model requires a potential function for each considered object type [105]. After aggregating the potentials functions one can obtain the maneuver model (MM) for a given starting point e.g. by applying gradient descent to the combined potential function.

*Stochastic reachable sets* Stochastic reachable sets are based on multi-trace branching prediction [5]. There are multiple approaches to using stochastic reachable sets in motion modeling, where the reachable sets arise from variation of the control inputs of the underlying single-trace dynamics model. When applying distributions to the variation of the control parameters and modeling the possible trajectories using Markov chains on a finite partitioning of the reachable sets, one obtains the most commonly used MM derived from stochastic reachable sets. Notice that this MM still depends on the underlying dynamics model used to obtain the reachable sets, thus it is always required to use another (nonlinear) MM together with this approach.

### 5.2 Criticality Metrics

Subsequently, we present a wide variety of criticality metrics, thereby delivering an extensive state of the art review. This analysis follows the approach described in Sect. 1.3. For each metric, we generally present a description alongside its corresponding formula.

In Sect. 1, criticality metrics have been introduced as a function $\kappa : \mathcal{S} \to \mathcal{O}$, with $\mathcal{S}$ the set of all traffic scenes and $\mathcal{O} \subseteq \mathbb{R} \cup \{-\infty, +\infty\}$ a measurement scale. Generally, criticality metrics can be differentiated into such *scene level* metrics—calculated given only the information at a specific point in time—and *scenario level* metrics—calculated for a given time series. Naturally, every scene level metric can be used to derive scenario level metrics by aggregation over time, but not vice versa. In the following descriptions, scene level metrics do have a point in time $t$ as an input (i.e. $\kappa(\cdot, t)$), while scenario level metrics do not (i.e. $\kappa(\cdot)$).

In order to avoid ambiguities, we note that the term 'metric' is used with various interpretations in the literature. Mathematically, the term is used for a function that measures the distance between two elements of a set. In the context of criticality metrics for automated vehicles, the term 'metric' is typically understood as a mathematical *measure* rather than a mathematical *metric*. The term 'metric' in this context originates from the engineering sciences, especially software engineering, where it has been defined as 'a measure of the extent [...] to which a product [...] possesses and exhibits certain [...] characteristics' [11].

*Aggregation over Time* While aggregation over time will be a non-issue for the remainder of this section, let us briefly elaborate. As already mentioned, any scene level criticality metric $\kappa(\cdot, t)$ can be used to define myriad scenario level metrics using suitable aggregate functions, i.e. $\kappa(\cdot, [t_0, t_e]) = \text{agg}_{t \in [t_0, t_e]} \kappa(\cdot, t)$. As evaluation of $\kappa(\cdot, t)$ on measurements along $[t_0, t_e]$ implies discretization, popular choices for aggregation function over discrete time include $\text{agg} \in \{\min, \max, \text{mean}, p\text{-quantile}, \text{median}, \sum, \ldots\}$. The choice of an appropriate aggregate is highly dependent on the criticality metric and the context of its application. The authors suggest that no universal statement about the 'optimal' aggregation over time can be made at this point and comparative experimental data is to be collected.

*Aggregation over Actors* Similarly, any criticality metric that is defined for one or two actors, e.g. $\kappa(A_1, A_2, \cdot)$, can naturally be extended to arbitrary actors in a scene or scenario by aggregation, e.g. $\kappa(\mathcal{A}, \cdot) = \text{agg}_{A_i, A_j \in \mathcal{A}, A_i \neq A_j} \kappa(A_i, A_j, \cdot)$. Depending on the actors of interest, a different aggregate might be appropriate: while for assessing the risk of a designated actor $A_1$, such as a vehicle of interest, aggregating $\kappa(A_1, \mathcal{A}, \cdot) = \max_{A_j \in \mathcal{A}, A_j \neq A_1} \kappa(A_1, A_j, \cdot)$ can be advantageous, a criticality assessment of a scene which is impartial to individual perspectives can benefit from aggregating $\kappa(\mathcal{A}, \cdot) = \text{mean}_{A_i, A_j \in \mathcal{A}, A_i \neq A_j} \kappa(A_i, A_j, \cdot)$. Analogously to aggregation over time, we exclude such considerations in this work, but acknowledge their necessity.

*Implementation and Computation* The focus of the subsequent review lies on presenting idealized formulae for each metric, stating constraints or assessing properties of interest for variables as estimated by a given prediction model. As an example, such an idealized formula can retrieve the future position of actor $A_i$ after a duration $\tilde{t}$ by using $\boldsymbol{p}_i(t + \tilde{t})$. Note that for an implementation of a scene level metric, one might want to limit the time horizon $t_H > 0$ of predictions used by the metric, i.e. constraining $\tilde{t} \in [0, t_H]$.

By using clear interfaces to the prediction models, dependencies thereon are both minimized and distinctly defined, conceptually enabling a substitution of various models implementing the employed interfaces. We hence refer to the prediction models for implementation details such as reaction times, actuator dynamics, discretization of time, and time horizons. Altogether, this enables the suitability analysis to abstract from such details, lifting its scope to the measurement principle instead of binding it to specific realizations. This abstraction ultimately allows to derive novel computational methods that implement the idealized representation as well as the comparison of such implementations to the measurement principle.

### 5.2.1 Time to Collision (TTC)

For two actors $A_1$, $A_2$ at time $t$, the TTC metric returns the minimal time until $A_1$ and $A_2$ collide according to a given DMM, or infinity if the predicted trajectories do not intersect [33, 94]. It is defined by

$$\text{TTC}(A_1, A_2, t) = \min\left(\{\tilde{t} \geq 0 \mid d(\boldsymbol{p_1}(t + \tilde{t}), \boldsymbol{p_2}(t + \tilde{t})) = 0\} \cup \{\infty\}\right).$$

A variety of the TTC, called Modified Time To Collision (MTTC), is extended under the name of Crash Index (CrI), where it is multiplied with a velocity-based severity estimate [74].

For car following scenarios and from the point of view of a distinguished actor, the TTC delivers a quality estimate on the temporal proximity to a collision that is induced by a maneuver of an actors, e.g. by a braking maneuvers of a lead vehicle. Its validity is however greatly reduced for most DMMs within intersection scenarios, cf. Fig. 1, as well as, if not meaningfully aggregated over actors, in multi actor scenes. Furthermore, the resulting time still needs to be interpreted w.r.t. the abilities and environment of $A_1$, either by using appropriate target values or composed metrics such as TTM.

One possible aggregate of the TTC to the scenario level is the Time To Accident (TTA) metric which is defined as $\text{TTA}(A_1, A_2) = \text{TTC}(A_1, A_2, t_{evasive})$ with $t_{evasive}$ being the first time where an evasive maneuver is performed [45]. Such aggregations over time can increase the TTC's validity when used for a retrospective assessment. Further information is given when discussing the other two time aggregates of TTC in this work, TET and TIT.

### 5.2.2 Potential Time to Collision (PTTC)

The PTTC metric, as proposed by Wakabayashi et al. [103], constraints the TTC metric by assuming constant velocity of $A_1$ and constant deceleration of $A_2$ in a car following scenario, where $A_1$ is following $A_2$. Given these constraints, the formula simplifies to

$$\text{PTTC}(A_1, A_2, t) = \frac{1}{a_{2,long}} \left( -\dot{d} \pm \sqrt{\dot{d}^2 + 2a_{2,long}d} \right)$$

with $\dot{d} = \dot{d}(\boldsymbol{p_1}(t), \boldsymbol{p_2}(t))$ and $d = d(\boldsymbol{p_1}(t), \boldsymbol{p_2}(t))$ respectively. While imposing such constraints on the scenario type and the DMMs of the actors reduces the computational cost of evaluating the metric, its validity is significantly reduced compared to the general TTC.

### 5.2.3 Time to Zebra (TTZ)

Defined by Várhelyi et al. [100], the TTZ measures the time until an actor $A_1$ reaches a zebra crossing CA, hence

$$\text{TTZ}(A_1, \text{CA}, t) =$$
$$\min \left( \{ \tilde{t} \geq 0 \mid d(\boldsymbol{p_1}(t+\tilde{t}), \boldsymbol{p_{\text{CA}}}(t+\tilde{t})) = 0 \} \cup \{\infty\} \right).$$

Note that this concept can be further generalized to a Time To Object (TTO) metric for arbitrary moving or non-moving objects and conflict areas. For moving objects, this generalization coincides with the TTC.

### 5.2.4 Time to Closest Encounter (TTCE)

The TTCE is a distance-dependent risk indicator, which generalizes the concept of the TTC to the non-collision case [20]. At time $t$, the TTCE returns the time $\tilde{t} \geq 0$ which minimizes the distance to another actor in the future. The corresponding minimal distance is called the Distance of Closest Encounter (DCE). The formulae are given as

$$\text{DCE}(A_1, A_2, t) = \min_{\tilde{t} \geq 0} d(\boldsymbol{p_1}(t+\tilde{t}), \boldsymbol{p_2}(t+\tilde{t})),$$
$$\text{TTCE}(A_1, A_2, t) = \arg \min_{\tilde{t} \geq 0} d(\boldsymbol{p_1}(t+\tilde{t}), \boldsymbol{p_2}(t+\tilde{t})).$$

In particular, as DCE $\to 0$, TTCE $\to$ TTC which implies that DCE $= 0$ if and only if TTCE $=$ TTC. Building on the TTCE and DCE, Eggert uses an exponential transform together with a survival function in order to estimate the future event probability of a collision for the distance-dependent risk [20].

### 5.2.5 Worst Time to Collision (WTTC)

Originally defined by Wachenfeld et al. [101], the WTTC metric extends the usual TTC by considering multiple traces of actors as predicted by an over-approximating DMM, i.e.

$$\text{WTTC}(A_1, A_2, t) = \min_{\boldsymbol{p_1} \in \text{Tr}_1(t), \boldsymbol{p_2} \in \text{Tr}_2(t)}$$
$$(\{\tilde{t} \geq 0 \mid d(\boldsymbol{p_1}(t+\tilde{t}), \boldsymbol{p_2}(t+\tilde{t})) = 0\} \cup \{\infty\}),$$

where $\text{Tr}_1(t)$ resp. $\text{Tr}_2(t)$ denotes the set of all possible trajectories available to actor $A_1$ resp. $A_2$ at time $t$, as constraint by the employed DMM. By design, the WTTC excels in selective data recording and data filtering applications.

### 5.2.6 Time Exposed TTC (TET)

The TET is a scenario level metric that builds on the TTC together with a target value $\tau$ [45, 66]. While originally defined only for discrete time, the formula can be elegantly generalized to continuous time as

$$\text{TET}(A_1, A_2, \tau) = \int_{t_0}^{t_e} \mathbf{1}_{\text{TTC}(A_1, A_2, t) \leq \tau} \, \text{dt}$$

where $\mathbf{1}$ denotes the indicator function. Therefore, TET measures the amount of time for which the TTC is below a given target value $\tau$. Its dependency on the scenario duration could easily be eliminated through division by $t_e - t_0$. Moreover, let us mention that the idea of 'time exposed below target value' can readily be adapted for any metric together with a target value and is essentially independent of the TTC.

### 5.2.7 Time Integrated TTC (TIT)

Similar to the TET, the Time Integrated TTC (TIT) [66] is a scenario level metric based on the TTC and is given as

$$\text{TIT}(A_1, A_2, \tau) =$$
$$\int_{t_0}^{t_e} \mathbf{1}_{\text{TTC}(A_1, A_2, t) \leq \tau} (\tau - \text{TTC}(A_1, A_2, t)) \text{dt}.$$

It aggregates the difference between the TTC and a target value $\tau$ in the time interval $[t_0, t_e]$. Therefore, the metric reflects criticality more accurately than the TET. As for the TET, the construction of the TIT can be adapted for other metrics.

### 5.2.8 Time to Maneuver (TTM)

The TTM metric [35, 94] returns the latest possible time in the interval [0, TTC] such that a considered maneuver performed by a distinguished actor $A_1$ leads to collision avoidance or $-\infty$, if a collision cannot be avoided. Therefore,

$$\text{TTM}(A_1, A_2, t, m) = \max \left( \{ \tilde{t} \in [0, \text{TTC}(A_1, A_2, t)] \mid \right.$$
$$\left. d(\boldsymbol{p_{1,m}}(t+s, t+\tilde{t}), \boldsymbol{p_2}(t+s)) > 0 \, \forall s \geq \tilde{t} \} \cup \{-\infty\} \right).$$

For analytic purposes, an extension of the TTM's output scale to negative values is possible. Various special cases of the TTM metric have been considered in the literature [48, 94, 102], including Time To Brake (TTB) [64] (i.e. $m$

= 'brake'), Time To Steer (TTS) [35] (i.e. $m$ = 'steer'), and Time To Kickdown (TTK) [35] (i.e. $m$ = 'kickdown').

### 5.2.9 Time to React (TTR)

The TTR metric [35, 94] approximates the latest time until a reaction is required by aggregating the maximum TTM metric over a predefined set of maneuvers $M$, i.e.

$$\text{TTR}(A_1, A_2, t) = \max_{m \in M} \text{TTM}(A_1, A_2, t, m).$$

For example, one might select $M$ = {'brake', 'steer', 'kickdown'}.

### 5.2.10 Time Headway (THW)

The THW metric calculates the time until actor $A_1$ reaches the position of a lead vehicle $A_2$ [43, 48], i.e.

$$\text{THW}(A_1, A_2, t) = \min\{\tilde{t} \geq 0 \mid \boldsymbol{p_1}(t + \tilde{t}) = \boldsymbol{p_2}(t)\}.$$

Analogously to the THW, one can define the Headway (HW) metric [43] simply as the distance to a lead vehicle, i.e.

$$\text{HW}(A_1, A_2, t) = d(\boldsymbol{p_1}(t), \boldsymbol{p_2}(t)).$$

The THW is used by regulatory bodies in several countries to express recommendations and as a threshold for fines [48].

### 5.2.11 Encroachment Time (ET)

The ET metric [4] measures the time that an actor $A_1$ takes to encroach a designated conflict area CA, i.e.

$$\text{ET}(A_1, \text{CA}) = t_{\text{exit}}(A_1, \text{CA}) - t_{\text{entry}}(A_1, \text{CA}).$$

While the value of ET is loosely correlated with criticality, it completely ignores the dynamics and behavior of other actors.

### 5.2.12 Post Encroachment Time (PET)

The PET [4] has been widely used as metric for the a-posteriori analysis of traffic data [58, 45, 75]. The PET calculates the time gap between one actor leaving and another actor entering a designated conflict area CA on scenario level. Assuming $A_1$ passes CA before $A_2$, the formula is

$$\text{PET}(A_1, A_2, \text{CA}) = t_{\text{entry}}(A_2, \text{CA}) - t_{\text{exit}}(A_1, \text{CA}).$$

Allen et al. also introduce two semi-predictive versions of the PET, called Gap Time (GT) and Initially Attempted Post Encroachment Time (IAPE), which inherit the properties of the PET and are not considered any further here [4]. Both metrics, GT and IAPE, measure $t_{\text{exit}}(A_1, \text{CA})$ and predict $t_{\text{entry}}(A_2, \text{CA})$ at different points in time using a constant velocity model. Therefore, they can be seen as an evaluation of the Predictive Encroachment Time (PrET) at a specific time point.

### 5.2.13 PrET (Predictive Encroachment Time)

Here, we summarize the predictive versions of the Post Encroachment Time. The redictive Encroachment Time (PrET) [69] is the anticipated PET w.r.t. a predicted intersection point:

$$\text{PrET}(A_1, A_2, t) = \min(\{|\tilde{t}_1 - \tilde{t}_2| \mid \boldsymbol{p_1}(t + \tilde{t}_1) = \boldsymbol{p_2}(t + \tilde{t}_2), \tilde{t}_1, \tilde{t}_2 \geq 0\} \cup \{\infty\}).$$

The Time Advantage (TA) metric [58] can be interpreted as a special case of PrET for a constant velocity model, i.e. $\boldsymbol{p}_i(s + t) = \boldsymbol{p}_i(t) + s\boldsymbol{v}_i(t)$. A scaled variant of the PrET, labeled Scaled Predictive Encroachment Time (SPrET), modifies the value of PrET by multiplication with the factor $(\tilde{t}_1 + \tilde{t}_2)$, i.e.

$$\text{SPrET}(A_1, A_2, t) = \\ \min(\{|\tilde{t}_1^2 - \tilde{t}_2^2| \mid \boldsymbol{p_1}(t + \tilde{t}_1) = \boldsymbol{p_2}(t + \tilde{t}_2), \tilde{t}_1, \tilde{t}_2 \geq 0\} \cup \{\infty\}),$$

in order to decrease the weight of situations long before the predicted intersection [69]. Therefore, the SPrET incorporates prediction uncertainty.

### 5.2.14 Proportion of Stopping Distance (PSD)

The PSD metric, proposed by Allen et al., is defined as the distance to a conflict area CA divided by the Minimum Stopping Distance (MSD) [4, 7, 30, 65]. Therefore,

$$\text{PSD}(A_1, \text{CA}, t) = \frac{d(\boldsymbol{p_1}(t), \boldsymbol{p}_{\text{CA}}(t))}{\text{MSD}(A_1, t)},$$

$$\text{MSD}(A_1, t) = \frac{v_{1,long}(t)^2}{2|a_{1,long,min}(t)|}.$$

### 5.2.15 Accepted Gap Size (AGS)

he AGS is a quantity which can be used to measure the complexity of a traffic situation. In general it quantifies the gap or the actual space between actors desired or required for others to make a positive action decision. Therefore, for an actor $A_1$ at time $t$, the AGS [62, 77] is the spatial distance that is predicted for $A_1$ to act, i.e.

$$\text{AGS}(A_1, t) = \min\{s \geq 0 \mid action(A_1, t, s) = 1\},$$

where a model $action(A_1, t, s)$ predicts, based on the circumstances at $t$, whether $A_1$ decides to act given the gap size $s$. This model can for example refer to the size of the gap in a stream of pedestrians passing a crosswalk, which is required for a waiting driver to decide to cut in and continue. For a time dependent distance measure, the metric is also called the accepted lag size. In general the more critical a traffic situation is, the larger the desired distance to other actors will be. For example, at an intersection, drivers tend to wait if the situation is unclear and the intersection itself is already crowded.

### 5.2.16 Required Longitudinal Acceleration ($a_{long,req}$)

For two actors $A_1, A_2$ at time $t$, $a_{long,req}$ measures the maximum longitudinal backward acceleration required, on average, by actor $A_1$ to avoid a collision in the future. It can be formalized as

$$a_{long,req}(A_1, A_2, t) = \max\{a_{1,long} \leq 0 \mid \forall \tilde{t} \geq 0 : d(\boldsymbol{p_1}(t+\tilde{t}), \boldsymbol{p_2}(t+\tilde{t})) > 0\}.$$

The $a_{long,req}$ can be adapted for the situation where the acceleration of $A_1$ needs to be positive in order to avoid a collision by taking the minimum $a_{1,long} \geq 0$ instead. An interesting special case is exhibited, when constant acceleration of the actors is assumed in a car following scenario, cf. [43, 5.3.5]. Assuming $A_1$ is following $A_2$, we have that

$$a_{long,req}(A_1, A_2, t) = \min\left(a_{2,long} + \frac{(v_{1,long}(t) - v_{2,long}(t))^2}{2d(\boldsymbol{p_1}(t), \boldsymbol{p_2}(t))}, 0\right).$$

Moreover, Jansson discusses the interesting cases of piecewise constant motion [43, 5.3.6] and the inclusion of actuator dynamics [43, 5.3.7]. For constant acceleration, the concept of $a_{long,req}$ is also known as Deceleration Rate to Avoid Crash (DRAC) [6].

### 5.2.17 Required Lateral Acceleration ($a_{lat,req}$)

Similar to the $a_{long,req}$, the $a_{lat,req}$ [43, 5.3.8] is defined as the minimal absolute lateral acceleration in either direction that is required for a steering maneuver to evade collision. For two actors $A_1, A_2$ at time $t$, $a_{lat,req}$ measures the minimum absolute lateral acceleration required, on average, by actor $A_1$ to avoid a collision in the future:

$$a_{lat,req}(A_1, A_2, t) = \min\{|a_{1,lat}| \mid \forall \tilde{t} \geq 0 : d(\boldsymbol{p_1}(t+\tilde{t}), \boldsymbol{p_2}(t+\tilde{t})) > 0\}.$$

For actors $A_1$ and $A_2$ with constant acceleration where $A_1$ is following $A_2$, the formula concretizes to

$$a_{lat,req}(A_1, A_2, t) = \min\{|a_{1,lat,left}(A_1, A_2, t)|, |a_{1,lat,right}(A_1, A_2, t)|\}$$

where

$$a_{1,lat,k}(A_1, A_2, t) = a_{2,lat,k} + \frac{2(v_{2,lat}(t) - v_{1,lat}(t))}{\text{TTC}(A_1, A_2, t)} + \frac{2}{\text{TTC}(A_1, A_2, t)^2} \cdot \left[\pm\left(\frac{w_1 + w_2}{2}\right) + p_{2,lat}(t) - p_{1,lat}(t)\right]$$

with $w_i$ denoting the width of $A_i$ and $k \in \{left, right\}$ depends on the sign of $\frac{w_1 + w_2}{2}$.

### 5.2.18 Required Acceleration ($a_{req}$)

Based on $a_{long,req}$ and $a_{lat,req}$, the aggregate metric $a_{req}$ can be defined in various ways [43, 5.3.10], e.g. by taking the norm of the required acceleration of both directions, i.e.

$$a_{req}(A_1, A_2, t) = \sqrt{a_{long,req}(A_1, A_2, t)^2 + a_{lat,req}(A_1, A_2, t)^2}.$$

More complex aggregates might also take into account the maximally available acceleration in each direction by incorporating the coefficient of friction $\mu$. Also, let us mention the Conditional Required Acceleration ($a_{req,cond}$) [69] which combines $a_{req}$ and SPrET for the analysis of urban intersection scenarios:

$$a_{req,cond}(A_1, A_2, t) = \begin{cases} a_{req}(A_1, A_2, t), & \text{if SPrET}(A_1, A_2, t) < 3s^2, \\ 0, & \text{otherwise.} \end{cases}$$

The $a_{req,cond}$ demonstrates by example how new criticality metrics can be created by combination of existing metrics and target values. In particular, the conditionality of the $a_{req,cond}$ encodes that the dynamical aspects of criticality only become relevant when a certain temporal criticality is present. This construction, of course, can be generalized as it is not specific to the $a_{req}$ and SPrET. Generally, addressing the different aspects of criticality through combination of metrics can lead to significant improvements in validity.

### 5.2.19 Deceleration to Safety Time (DST)

For an actor $A_1$ following another actor $A_2$, the DST metric calculates the deceleration (i.e. negative acceleration) required by $A_1$ in order to maintain a *safety time* of $t_s \geq 0$ seconds under the assumption of constant velocity $v_2$ of actor $A_2$ [38, 87]. The corresponding formula can be written as

$$\text{DST}(A_1, A_2, t, t_s) = \frac{(v_{1,long}(t) - v_{2,long}(t))^2}{2(d(\boldsymbol{p_1}(t), \boldsymbol{p_2}(t)) - v_{2,long}(t) \cdot t_s)}$$

and extends the concept of the $a_{long,req}$ by requiring deceleration to a safety distance $v_{2,long}(t) \cdot t_s$, under the assumptions of constant velocity of $A_2$. In particular, for $t_s = 0$, the DST agrees with the constant acceleration version of $a_{long,req}$.

### 5.2.20 Brake Threat Number (BTN)

For actor $A_1$, the BTN [43] is defined as the required longitudinal acceleration imposed on actor $A_1$ by actor $A_2$ at time $t$, divided by the longitudinal acceleration that is at most available to $A_1$ in that scene, i.e.

$$\text{BTN}(A_1, A_2, t) = \frac{a_{long,req}(A_1, A_2, t)}{a_{1,long,min}(t)}.$$

By definition, a BTN $\geq 1$ indicates that a braking maneuver performed by $A_1$ cannot avoid an impeding accident under the assumed DMM. An extension of BTN to multiple actors is proposed by Eidehall [21]. For car following scenarios, a special case of the BTN, known as the Deceleration-based Surrogate Safety Measure (DSSM), incorporates human factors into the model by combining the worst-case assumption of $A_2$ braking maximally with $A_1$'s required reaction time [92].

### 5.2.21 Steer Threat Number (STN)

Similar to the BTN, for two actors at time $t$, the STN [21, 43] is defined as the required lateral acceleration divided by the lateral acceleration at most available to $A_1$ in that direction:

$$\text{STN}(A_1, A_2, t) = \frac{a_{lat,req}(A_1, A_2, t)}{a_{1,lat,max}(t)}.$$

By definition, an STN $\geq 1$ indicates that a lateral maneuver performed by $A_1$ cannot avoid an impeding accident.

### 5.2.22 Lateral Jerk (LatJ) and Longitudinal Jerk (LongJ)

Jerk, being the rate of change in acceleration and thus quantifying the abruptness of a maneuver, is formulated as

$$\text{LatJ}(A_1, t) = j_{1,lat}(t), \quad \text{LongJ}(A_1, t) = j_{1,long}(t).$$

The jerk has many immediate applications such as discerning different classes of driving styles, e.g. comfortable, angry, anxious, and risky modes [10, 24]. Another important application area are trains and buses, where for standing passengers, the jerk enables an analysis of their reaction capabilities on the maneuver, e.g. during a change of tracks of a train [78].

The usage of LongJ and LatJ varies, e.g. LongJ can be utilized in the design of an adaptive cruise control (ACC) [41] function, whereas LatJ is used for functions dealing with steering maneuvers, e.g. a Lane Keeping Assistance System (LKAS) [40].

### 5.2.23 Space Occupancy Index (SOI)

The SOI defines a personal space for a given actor actor and counts violations by other participants while setting them in relation to the analyzed period of time [45, 72, 95]. For each actor $A_i$ at time $t$, a personal space $Sp(A_i, t)$ is defined. At time $t$, if there exists some $A_j \neq A_i$ s.t. $Sp(A_i, t) \cap Sp(A_j, t) \neq \emptyset$, a violation of the personal space of $A_i$ is given. The number of conflicts is then given as $C(A_1, \mathcal{A}, t) = \sum_{A_j \in \mathcal{A} \setminus \{A_1\}} [Sp(A_1) \cap Sp(A_j) \neq \emptyset]$, where $[\cdot]$ denotes the Iverson bracket. Thus, for a given scenario in the time interval $[t_0, t_e]$, the conflict index is defined as

$$\text{SOI}(A_1, \mathcal{A}) = \sum_{t=t_0}^{t_e} C(A_1, \mathcal{A}, t).$$

SOI was introduced for bicycles and pedestrians, however, it is possible to formulate a similar concept for road vehicles.

### 5.2.24 Pedestrian Risk Index (PRI)

The PRI estimates the conflict probability and severity for pedestrian crossing scenarios by combining the TTZ with the impact speed [15]. It is defined for a scenario with a vehicle $A_1$ and a vulnerable road user (VRU) $P$ both approaching a conflict area CA. The scenario shall include a unique and coherent conflict period $[t_{c_{start}}, t_{c_{stop}}]$ where $\forall t \in [t_{c_{start}}, t_{c_{stop}}] : \text{TTZ}(P, \text{CA}, t) < \text{TTZ}(A_1, \text{CA}, t) < t_s(A_1, t)$. Here, $t_s(A_1, t)$ is the time $A_1$ needs to come to a full stop at time $t$, including its reaction time, leading to

$$\text{PRI}(A_1, \text{CA}) = \int_{t_{c_{start}}}^{t_{c_{stop}}} (s_{imp}(A_1, \text{CA}, t)^2 \cdot (t_s(A_1, t) - \text{TTZ}(A_1, \text{CA}, t))) \text{dt},$$

where $s_{imp}$ is the predicted speed at the time of contact with the pedestrian crossing. The PRI thus quantifies over two aspects of a whole scenario: the temporal difference is claimed to be a surrogate for the accident probability, whereas the impact speed is approximate for its severity. One possibility of estimating $s_{imp}$ is defined by the authors as

$$s_{imp}(A_1, \text{CA}, t) = \\ \sqrt{\|\mathbf{v_1}(t)\|_2^2 + 2a_{1,long,min}(t)(d(\mathbf{p_1}(t), \mathbf{p}_{\text{CA}}(t)) - \|\mathbf{v_1}(t)\|_2 t_1^r)},$$

where $t_i^r$ is the reaction time of actor $A_i$. Note that depending on the DMM, other formulae for $s_{imp}$ may be employed.

### 5.2.25 Crash Potential Index (CPI)

The CPI is a scenario level metric and calculates the average probability that a vehicle can not avoid a collision by deceleration. It sums over the probabilities that a given vehicle's $a_{long,req}$ exceeds its $a_{long,min}$ for each time point and normalizes the value over the length of the scenario [18]. The target value $a_{long,min}$ is assumed to be normally distributed and dependent on factors such as road surface material and vehicle brakes. While originally defined in discrete time, the CPI can be defined in continuous time as:

$$CPI(A_1, A_2) = \frac{1}{t_e - t_0} \int_{t_0}^{t_e} P(a_{long,req}(A_1, A_2, t) < a_{1,long,min}(t)) \mathrm{d}t .$$

Note that this concept of aggregation over time can be generalized to be applicable to other metrics, assuming that a valid probability distribution of the target value can be given. This potentially enables a more precise identification of criticality within a scenario.

### 5.2.26 Aggregated Crash Index (ACI)

The ACI measures the collision risk for car following scenarios by extending the concept of CPI from $a_{req}$ to multiple conditions. First, a probabilistic causal model of the scenario type under consideration is needed to derive a collision tree with all possible outcomes and their probabilities [56].

The concrete outcomes are represented by the tree's leaf nodes $L_j$. Every leaf node has a value $C_{L_j}$ which is 0 in case of no collision and 1 in case of a collision. Similar to CPI, the conditions are defined based on other metrics, e.g. the current stopping time of the lead vehicle being smaller than a lognormally distributed reaction time. The collision risk $CR_{L_j}(S, t)$ of a leaf node $L_j$ given a scene $S$ at time $t$ is hence represented by $CR_{L_j}(S, t) = P(L_j, t) \cdot C_{L_j}$, where $P(L_j, t)$ is the probability of satisfying all conditions necessary to reach $L_j$, when given the conditions at time $t$. Finally, the ACI is an aggregation of the collision risks in a scene $S$ at time $t$:

$$\text{ACI}(S, t) = \sum_{j=1}^{n} CR_{L_j}(S, t),$$

with $n$ being the number of leaf nodes in the collision tree.

### 5.2.27 Trajectory Criticality Index (TCI)

The TCI metric models criticality using an optimization problem [49]. The task is to find a minimum difficulty value, i.e. how demanding even the easiest option for the vehicle will be under a set of physical and regulatory constraints. For example, if the constraint is to avoid obstacles, then driving straight towards an obstacle and being only a few seconds away requires a large change in steering angle and acceleration to satisfy the constraint of collision avoidance.

Here, the possible set of vehicle actions are not only constrained by physically possible behavior; it additionally shall adhere to a mathematically modeled set of requirements. Said constraints are based on the necessary longitudinal ($a_{long}$) and lateral acceleration ($a_{lat}$) to avoid collisions as well as the margin for corrections in speed ($R_{long}$) and course angle ($R_{lat}$). Since both $R_{long}$ and $R_{lat}$ are dependent on $a_{long}$ and $a_{lat}$, it suffices to minimize the combined function w.r.t. $a_{long}$ and $a_{lat}$. The requirements include concepts such as holding a safe following distance and maximizing distance to obstacles.

Assuming the vehicle behaves according to Kamm's circle, TCI for a scene $S$ with an ego vehicle $A_1$ reads as

$$\text{TCI}(A_1, S, t, t_H) = \min_{a_{long}, a_{lat}} \sum_{\tilde{t}=t}^{t+t_H} w_{long} R_{long}(\tilde{t}) + w_y R_{lat}^2(\tilde{t}) \\ + \frac{w_{ax} a_{long}^2(\tilde{t}) + w_{ay} a_{lat}^2(\tilde{t})}{(\mu_{max} g)^2}$$

where $t_H$ is the prediction horizon, $a_x$ and $a_y$ the longitudinal and lateral accelerations, $\mu_{max}$ the maximum coefficient of friction, $g$ the gravitational constant, $w$ weights, and $R_{long}$ and $R_{lat}$ the longitudinal and lateral margin for angle corrections:

$$R_{long}(t) = \frac{\max(0, x(t) - r_{long}(t))}{d_{long}(t)},$$

$$R_{lat}^2(t) = \frac{(y(t) - r_{lat}(t))^2 v(t - t_s)}{d_{lat}^2(t) v_{max}}.$$

Here, $x(t), y(t)$ is the position, $t_s$ the discrete time step size, $v_{max}$ the maximum velocity, $r_{long}(t)$ the reference for a following distance (set to $2S \cdot v_{long}(t)$), $r_{lat}$ the position with the maximum lateral distance to all obstacles in $S$, $d_{long}(t)$, $d_{lat}(t)$ the maximum longitudinal and lateral deviations from $r_{long}, r_{lat}$.

### 5.2.28 Conflict Index (CI)

The conflict index enhances the PET metric with a collision probability estimation as well as a severity factor [3]. For this, the estimated kinetic energy that would have been released assuming a hypothetical collision between $A_1$ and $A_2$ at their states when entering ($A_2$) resp. exiting ($A_1$) the conflict area is estimated:

$$\text{CI}(A_1, A_2, \text{CA}, \alpha, \beta) = \frac{\alpha \Delta K_e}{e^{\beta \text{PET}_{(A_1, A_2, \text{CA})}}},$$

where the denominator is a collision probability estimation.

Therefore, it is proposed that the actual collision probability is proportional to $e^{-\beta \text{PET}_{(A_1, A_2, \text{CA})}}$ with $\beta$ being a calibration factor dependent on e.g. country, road geometry, or visibility, and $[\beta] = S^{-1}$. The nominator represents the collision severity, where $\alpha \in [0, 1]$ is again a calibration factor for the proportion of energy that is transferred from the vehicle's body to its passengers and $\Delta K_e$ is the predicted absolute change in kinetic energy before and after the predicted collision.

$\Delta K_e$ is estimated based on the masses as well as velocities and angles at time of entering ($A_2$) resp. exiting ($A_1$) CA.

### 5.2.29 Collision Probability via Monte Carlo (P-MC)

P-MC produces a collision probability estimation based on future evolutions from a Monte Carlo path planning prediction [12]. At first, a binary representation of the road geometry with the distinction of drivable and non-drivable is generated. If the ego enters a non-drivable region, a collision is detected. Every object in the scene has a state, denoted by $s_i(t) = (p_i(t), v_i(t))$, and control inputs $u_i(t)$. The motion of each object is then described by an ODE of the form $\dot{s}_i(t) = f(s_i(t), u_i(t))$.

If the bounding boxes of two objects intersect at some point between $t$ and $t + t_H$, a collision is detected. A goal function $g(u_i(t))$ is defined for each object in the scene to specify the desirability of paths that the object might follow based on the possible control inputs. With $k$ objects in a scene, the combined goal of all objects is defined as

$$P(\mathcal{U}) := P(u_1, \ldots, u_k) := \prod_{j=1}^{k} P(u_j)^{\alpha_j}.$$

For an actor $A_1$ in a scene $S$, the collision probability is then

$$\text{P-MC}(A_1, S, t) = P(\mathcal{C}) = \int P(\mathcal{C} \mid \mathcal{U}) P(\mathcal{U}) d\mathcal{U},$$

with $P(\mathcal{C} \mid \mathcal{U})$ being the collision probability of $A_1$ in $S$ under the given inputs $\mathcal{U}$.

### 5.2.30 Collision Probability via Stochastic Reachable Sets (P-SRS)

Althoff et al. propose to estimate a collision probability using stochastic reachable sets [5]. Firstly, the reachable set $R([t, t + t_H])$ (the set of possible positions within $t_H$) is over-approximated for each actor, where the movement of the actor is approximated by Markov chains with time steps $\{t + t_1, t + t_2, \ldots, t + t_H\}$ and a constant $T = t_{k+1} - t_k$. The ego's motion is not modeled as it is assumed to be known.

Afterwards, the state and input space are discretized, thus we can write $R_i^\alpha(T)$ for the reachable set given a state in the $i$-th partition of the state space and the input in the $\alpha$-th partition of the input space for time $T$. The transition probabilities to partitions $X_j$ of the state space are given by

$$\Phi_{ji}^\alpha(T) = \frac{V(R_i^\alpha(T) \cap X_j)}{V(R_i^\alpha(T))}$$

where $V$ returns the volume. Aforementioned concepts are then generalized to $\Phi_{ji}^\alpha([0, T])$ by taking the union of $R_i^\alpha(t)$ for $t \in [0, T]$, not accounting for the discrete time aspect at this point [5].

Using the properties of Markov chains, one can thus derive the probability distribution of the position for each time interval. Behaviors of other actors are modeled as Markov chains on the control input space of the DMMs. Due to the discretization of the state space, we can approximate the lateral deviation by a piecewise constant function and thus we can define intervals $D_f$ where said function is constant. This leads to a lateral position probability of

$$p_f^{dev}([t_k, t_{k+1}]) = P(\delta \in D_f, t \in [t_k, t_{k+1}]).$$

By splitting the state space partitions $X_i$ into position and velocity, i.e. $X_i = S_e \times V_m$, one can define

$$p_e^{path}([t_k, t_{k+1}]) = \sum_m P(s \in S_e, v \in V_m, t \in [t_k, t_{k+1}]).$$

Afterwards, all possible paths in which two actors could have intersecting vehicle bodies are identified and stored in a list $\Omega$. Under the assumption of stochastic independence and using the previous concepts, we then have $p_{ef}^{pos} = p_e^{path} \cdot p_f^{dev}$, hence leading to the collision probability

$$\text{P-SRS}(A_1, S, t) = p^{col} = \sum_{(g,h,e,f) \in \Omega} \hat{p}_{gh}^{pos} \cdot p_{ef}^{pos}.$$

### 5.2.31 Collision Probability via Scoring Multiple Hypotheses (P-SMH)

Similar to other probability-based approaches, Sánchez Morales et al. propose to assign probabilities to predicted trajectories and accumulate them into a collision probability [67]. The ego's motion is modeled by a two track model, cf. Sect. 5.1. Due to less information being known for the other actors, a one track model is used for those. Pedestrians have the ability of changing direction, velocity, and acceleration in a finite set of steps under given constraints. Once the number $N$ of trajectories for the ego and total number $M$ of trajectories of all other actors is determined, one can compute the collision probability of $A_1$ at time $t$ as

$$\text{P-SMH}(A_1, \mathcal{A}, t) = \sum_{i=1}^{N} \sum_{j=1}^{M} \chi_j^i p_{A_1,i} p_{(\mathcal{A} \setminus A_1),j},$$

where $\chi_j^i$ equals one if and only if the $i$-th trajectory of $A_1$ and the $j$-th trajectory of the actors in $\mathcal{A} \setminus A_1$ lead to a collision, and $p_{A_1,i}$ resp. $p_{(\mathcal{A} \setminus A_1),j}$ are the probabilities of the trajectories being realized.

### 5.2.32 Potential Functions as Superposition of Scoring Functions (PF)

The general concept of the PF metric is to define a potential function for each static or dynamic object considered by the metric [105]. This includes potentials for lane markings, the road geometry, other vehicles, or, in more urban areas, pedestrians and bicyclists. Once a potential function for each object in the scene, denoted by $U_i(A, S)$, is chosen, one can apply e.g. gradient descent for a given scene $S$ to the combined potential function $U(A, S) = U_1(A, S) + \cdots + U_k(A, S)$, where $A$ is an actor and $k$ denotes the number of objects. A simple example of how to evaluate this metric for an actor $A_1$ and a given scene $S'$ is by inserting the values into $U$, i.e.

$$\text{PF}(A_1, S') = U(A_1, S') = U_1(A_1, S') + \cdots + U_k(A_1, S').$$

However, methods involving the mentioned gradient descent to assess the criticality can improve precision and also provide a suggestion for criticality-reducing vehicle movement.

Due to the way this metric is defined, almost all properties depend on the specified potential functions. Furthermore, while ethical questions play a role when defining any safety surrogate, it becomes more evident for potential functions, as an active decision making in the definition of the potentials is required.

### 5.2.33 Safety Potential (SP)

Conceptually, the Safety Force Field (SFF) framework identifies, under the assumption of all actors conducting some safe control policy (e.g. an emergency brake), whether there can exist a conflict [70]. To measure safety w.r.t collision avoidance, SFF uses SP as a numeric valuation.

Formally, each safe control policy $s \in S_1$ brings an actor $A_1$ to a full stop in finite time. SFF defines the occupied set $O_1$ of an actor $A_1$ to include its safety margin as well as $A_1$ itself. For each point on each trajectory that can arise from conducting a safe control policy $s \in S_1$, $O_1$ is examined. The resulting union of trajectories is the claimed set $C_1$.

The unsafe set between two actors $A_1, A_2 \in \mathcal{A}$ can then be identified as $U_{1,2} = \{x = (x_1, x_2) \in \Omega_1 \times \Omega_2 \mid C_1(x_1) \cap C_2(x_2) \neq \emptyset\}$. Intuitively, it is the set of all actor state combinations for which there exist safe control policies leading to a collision.

Identifying the combined state space of $A_1$ and $A_2$ as $\Omega_1 \times \Omega_2$, SFF subsequently employs a potential function $\rho_{1,2} : \Omega_1 \times \Omega_2 \to \mathbb{R}$ to rate the combined states of actors, where

- $\rho_{1,2}(x) > 0$ for all $x \in U_{1,2}$ and
- $\rho_{1,2}(x) \geq 0$ for all $x \notin U_{1,2}$ and
- $\rho_{1,2}(x) \geq \rho_{1,2}(x')$ for all states $x'$ that arise from $x$ when $A_1$ and $A_2$ apply safety procedures $s_1, s_2 \in S_1, S_2$.

The safety potential can hence rate a two-actor scene from one of their perspectives.

The authors state the following exemplary safety potential for some $k \in \mathbb{Z}_{>0} \cup \{\infty\}$:

$$\text{SP}(A_1, A_2, t) = \rho_{1,2} = \|(t_{stop}(A_1) - t_{int}, t_{stop}(A_2) - t_{int})\|_k$$

where $t_{int}$ is the the earliest intersection time when continuing the current situation under some model, and $t_{stop}(A_i)$ is the time of full stop of $A_i$ after applying a safety procedure.

### 5.2.34 Accident Metric (AM)

AM evaluates whether an accident happened in a scenario:

$$\text{AM}(\text{Sc}) = \begin{cases} 0, & \text{no accident happened during Sc,} \\ 1, & \text{otherwise.} \end{cases}$$

This simplistic metric is implicitly used in accident databases, e.g. GIDAS[4]. It fails to identify critical non-accident scenarios.

### 5.2.35 Responsibility Sensitive Safety Dangerous Situation (RSS-DS)

The RSS framework is designed to formally guarantee safety during an automated vehicle's drive [89].

---

[4] www.gidas.org

For this, the safe lateral and longitudinal distances $d_{min}^{lat}$ and $d_{min}^{long}$ are formalized, depending on the current road geometry. The metric RSS-DS for the identification of a dangerous situation is therefore defined as

$$\text{RSS-DS}(A_1, \mathcal{A}) = \begin{cases} 1, \exists A_i \in \mathcal{A} \setminus \{A_1\}. \\ d^{lat}(A_1, A_i) < d_{min}^{lat} \wedge \\ d^{long}(A_1, A_i) < d_{min}^{long}, \\ 0, \text{otherwise.} \end{cases}$$

In order to determine $d_{min}^{lat}$ and $d_{min}^{long}$, different prediction models are used, e.g. for intersections, highways, and unstructured roads.

Note that RSS has been shown to not consider certain edge cases, e.g. during braking maneuvers and on varying road surfaces and slopes, as well as the issue of perception uncertainty [52].

An extension of RSS-DS measures the temporal extent to which the ego was not able to mitigate the dangerous situation [44]. In accident research, a similar concept of classifying situations as safe and unsafe depending on longitudinal stopping distances was introduced as the Stopping Distance Index (SDI) [73]. In turn, the SDI is partially based on the idea of the Potential Index for Collision with Urgent Deceleration (PICUD) [98], both comparing the stopping distances of the lead and following vehicle under emergency braking.

### 5.2.36 Delta-v (Δv)

$\Delta v$ is the change in speed over collision duration and widely used in collision databases, where it is typically calculated from post-collision measurements [25]. Introduced in the 1970s [16], it uses the difference in speed to estimate the probability of a severe injury or fatality:

$$\Delta v(A_1) = \|v_1(t_{aftercol})\|_2 - \|v_1(t_{beforecol})\|_2.$$

A more complex formula for two actors taking the masses into account is given by

$$\Delta v(A_1, A_2, t) = \frac{m_2}{m_1 + m_2} (\|v_2(t)\|_2 - \|v_1(t)\|_2).$$

An extended $\Delta v$ measure, which is additionally considering the mass as well as the driving angles of the collision participants, has been discussed by Laureshyn et al. [60].

Joksch [46] presents a model connecting $\Delta v$ to the probability $P$ of a two vehicle collision leading to a fatality using

$$P(A_1) \approx \left(\frac{\Delta v}{31.74 \text{ms}^{-1}}\right)^4.$$

This connection provides an easily interpretable measure.

### 5.2.37 Conflict Severity (CS)

CS is concerned with solely estimating the severity of a potential collision in a scenario [8]. It thus presents as a suitable factor that can enhance various collision probability metrics in ensuring a more accurate representation of criticality. From the perspective of an actor $A_1$ performing a braking maneuver at time $t_{evasive}$, it is defined as

$$\text{CS}(A_1, A_2) = \Delta v(A_1, A_2, t_{evasive}) - \left(\text{TTA}(A_1, A_2) \cdot \|a_1(t_{evasive})\|_2 \cdot \frac{m_2}{m_1 + m_2}\right).$$

Thus, it compares the (extended) $\Delta v$ at time of the evasive maneuver against the $\Delta v$ at the potential collision point as predicted by TTA if $A_1$ conducts an emergency braking, assuming $v_2(t_{evasive} + \text{TTA}(A_1, A_2)) = 0$. CS factors in the relative mass difference due to the correlation between severe injuries and fatality outcome, measured on the Abbreviated Injury Scale, and the mass ratio of the involved actors [22].

## 5.3 Properties of Criticality Metrics

### 5.3.1 Property assessment

In Sect. 4, we have presented a list of relevant properties of criticality metrics. Based on the previous depiction of the metrics, we can now perform an evaluation of their properties.

To this end, the authors have created a supplementary web page[5] on which a detailed assessment for each of the aforementioned metrics can be found. The analysis is based on the authors' hypotheses, which are, wherever possible, covered by evidences. We remark that for many metrics no suitable sources providing evidences for the properties are available in the literature, strongly suggesting further empirical research. The underlying repository is therefore open to enhancements and suggestions by the scientific and industrial community[6].

For brevity, we refrain from including the complete set of properties within this publication, although they will be referenced and used in Sect. 6.2. However, to foster comprehension of the performed property assessment, we subsequently discuss the properties of the TTB metric, as taken from the referenced web page[7], exemplarily in more depth.

---

[5] http://purl.org/criticality-metrics

[6] http://purl.org/criticality-metrics/repository

[7] http://purl.org/criticality-metrics/ttb

### 5.3.2 Properties of the TTB metric

*Run-time capability* The TTB metric is calculated given a scene $S$ at a specific time point $t$. Semantically, it predicts the latest time for which a brake maneuver can successfully mitigate a collision, therefore forecasting a future evolution of $S$. Factually, these two properties are the foundation of any criticality metric to be used in an online setting, e.g. within an ADF implementation.

*Target values* Target values give meaning to the metric's result within a specific application, e.g. enabling to classify scenes or scenarios. The literature suggests:

1. for pruning of an MRM for an ADF within an exemplary situation, a TTB $\geq$ 0.4s was used to classify braking as a viable MRM [94] (application A.3),
2. for scenario classification of car following and lane merging maneuvers, 1s was used as an estimate based on the human reaction time, while 0.6s was empirically observed as critical by human subjects [48] (application B.2),
3. for criticality-based scenario classification within simulation runs, 1s was used (in combination with other criticality metrics) [36] (application B.2).

Note that target values are to be considered within their context. For example, a target value of 0.4s may be very well used for pruning a set of MRMs, but is potentially unfit for filtering large data bases due to a low sensitivity.

*Subject type* The TTB requires that an MM for braking can be formalized for the subject of the metric ($A_1$). This is achievable for road vehicles, whether human- or automation-driven. For VRUs, such as bicyclists and pedestrians, a braking maneuver is complex to model, and often not the sole reaction to a critical situation, e.g. combined with steering. Therefore, the TTB metric is primarily concerned with road vehicles. Note that the type of $A_2$ is not constrained.

*Scenario type* If evaluated on a scene, the TTB requires a collision in the prediction model, i.e. TTC $\neq \infty$, to return a meaningful value. Otherwise, when the DMM does not predict a collision, the TTB yields $\infty$ even if the criticality is heightened. Consequently, if evaluated on a scenario, the TTB needs a significant time span in which a collision is reliably predicted within the employed DMM for the metric to return a time series of meaningful (i.e. non-infinite) valuations.

*Inputs* The TTB metric uses the current information on both actors $A_1$ and $A_2$ to predict future behaviors. It therefore requires their states—e.g. pose and shape—at the time of evaluation $t$. Moreover, as motivated earlier, an MM for a braking maneuver of $A_1$ is needed.

*Output scale* If a collision can not be avoided at any future time point, the TTB yields $-\infty$. Otherwise, a value in $[0, \text{TTC}(A_1, A_2, t)]$ is returned, which leads to the output scale of $\{-\infty\} \cup [0, \infty]$ having time as its quantity. As there exists an absolute point of zero, it is given on a ratio scale, therefore supporting proportional comparisons between values.

*Reliability* The subsequent properties rely heavily on the employed prediction model. Assume that the DMM predicts a collision at time $t$ in the near future. In that case, the TTB adequately reflects criticality w.r.t. a braking maneuver. Assume now that in the next time point $t' > t$, the DMM is not able to predict a collision anymore, e.g. due to a small change in yaw angle by $A_2$. This effectively leads to a TTB value of $\infty$. The actual criticality at $t'$ to $t$ may only be changed slightly, but the metric jumps from indicating a certain criticality level at $t$ to assessing the scene as uncritical at $t'$. Therefore, the TTB's reliability is only high under the assumption that collisions can be reliably predicted.

*Validity* The TTB's validity primarily depends on the validity of the collision predicted within the DMM. If the DMM assumes e.g. unrealistic motions, the TTB's validity will suffer. Furthermore, in case the TTB is evaluated solely under a fixed actor's perspective within scenes involving multiple actors, validity can be reduced. This can be mitigated by aggregating over all actors. For at least one, a braking maneuver will be indicated.

*Sensitivity* Both sensitivity and specificity are derived properties—they base on the validity of the metric, the target values, and the DMM. For example, the TTB's specificity may be reduced if no collisions are predicted for critical scenarios.

*Specificity* As braking is a key choice in human reaction to hazards [2], the TTB indicates the remaining time until no (realistic) mitigation maneuver is probable with a high specificity for humans. AVs may exhibit different abilities for mitigation maneuvers, e.g. by avoiding collisions through last-second steering, options—even if possible—often not conducted by human drivers [2]. Therefore, TTB's specificity may be reduced when used for AVs, as non-braking maneuvers can still avoid critical situations in which TTB $\leq 0$.

*Prediction model* The TTB uses the function $p_i : \mathbb{R}^+ \to \mathbb{R}^3$ as an interface to the DMM in order to retrieve a prediction on the positions of the actor $A_i$ at future time points. Therefore, the time window is theoretically unbound, but larger time horizons may decrease the prediction validity. Additionally, it considers only linear time as, for each time point, only a single future position is retrieved from $p_i$.

### 5.4 Interrelations of Criticality Metrics

This section detailed a vast set of metrics and their interrelations. For example, we showed that the BTN is dependent on the $a_{long,req}$ metric. Such interrelations are helpful to understand the complex network spanned by the metrics during the execution of a suitability analysis. Figure 5 gives an overview of those relations, where we differentiate between scenario and scene level metrics, as introduced in Sect. 5.2. Furthermore, metrics that do not rely on a prediction model are highlighted.

**Fig. 5** Visualization of the interrelations of criticality metrics as presented in Sect. 5.2

## 6 Suitability Analysis

Up to this point, we have extensively reviewed the state of the art of criticality metrics within their contexts. For an in-depth interpretation of these results, we propose a methodical *suitability analysis*, which can be understood as a synthesis of the preceding sections. Given an application (Sect. 3) and its associated requirements on properties (Sect. 4), the goal of the suitability analysis is to find a set of adequate metrics and models (Sect. 5). The suitability analysis can additionally be used to discard metrics that are, regardless of the models, *not* suitable for the considered application. We first sketch the generic approach to a suitability analysis and subsequently give an evaluation using a comprehensible example.

### 6.1 Suitability Analysis Process

The suggested suitability analysis is presented as a five-step, expert-based process. It is given as *inputs*:

– a description of the application at hand, $A$,
– a set of available metrics, $K$,
– and a set of available models, $M$.

Its *output* is defined to be a set of suitable metric that are assigned a set of suitable models, $\mathcal{K} \subseteq K \times 2^M$. The work flow of the process takes the following shape:

(1) Identify a finite set of properties $P$ of criticality metrics relevant for $A$. Derive a set of requirements $R$ on the properties, potentially based on Table 4.
(2) Order the requirements $R$ w.r.t. their importance for the application $A$. This results in a relation $(R, \geq)$.
(3) For each available metric with its potential models, $(\kappa, \eta_\kappa) \in K \times 2^M$, determine its properties, leading to an assessment similar to Sect. 5.3, tailored to the details of $A$, relevant properties $R$, and available models $M$.
(4) Choose some $r \in \max(R)$, discard all metrics that do not fit the requirements from (3), and remove $r$, i.e.
1. $\mathcal{K} := \mathcal{K} \setminus \{(\kappa, \eta_\kappa) \in \mathcal{K} \mid (\kappa, \eta_\kappa) \not\models r\}$,
2. $R := R \setminus \{r\}$.
(5) Repeat (4) until either
1. $|R| = 0$ and $|\mathcal{K}| \geq 1$, then return $\mathcal{K}$.
2. $|R| \neq 0$ and $|\mathcal{K}| = 0$, then either conclude that there is no suitable metric for $A$ or restart the process with

– a relaxed set of requirements $R$, or
– an enlarged set of metrics and models $\mathcal{K}$.

The semantics of the relation $(\kappa, \eta_\kappa) \models r$ can be understood as the satisfaction of the requirement $r$ on the corresponding property $p \in P$ of $\kappa$ for all models in $\eta_\kappa$. It can only be detailed on a per-application basis and is typically not formally evaluated but rather inspected by an application expert.

### 6.2 Evaluation of the Suitability Analysis

As to present a concise evaluation of the suggested suitability analysis, we apply the process delineated in Sect. 6.1 to a concretely specified application, demonstrating the relevancy and value of the state of the art review of Sects. 3, 4, and 5.

#### 6.2.1 Exemplary Application

In this application, a prior analysis identified the phenomenon 'unprotected left turn' as highly relevant as well as a scenario class featuring the phenomenon. This class is to be analyzed w.r.t. criticality and the results are to be used within a scenario-based testing approach for an AV [68]. In our example, only limited resources are available. Hence, for metrics with an unconstrained DMMs, it is preferred to use a physics-based model as presented in Sect. 5.1.

The application generates critical instances from a logical scenario with two parameters, with the initial scene given in Fig. 6. It contains two actors on a four-arm intersection, with $A_1$ on the southern arm intending a left turn and $A_2$ on the northern arm intending a straight passing. We assume a mixed traffic environment, i.e. $A_2$ can be a human driver. The application's procedure for scenario instantiation is:

1. Using a state of the art simulation environment, sample uniformly from the logical scenario space $[0, 40] \times [5, 50]$, carry out the simulation, and compute a criticality measurement for each sample.
2. Fit a suitable regression model on the generated data using the measured criticality as the dependent variable.
3. Use an optimization algorithm on the learned model to identify the critical parameter combinations.

The identified critical parameter combinations are then used in a downstream testing process as a representatives for the phenomenon 'unprotected left turn' in the associated scenario.

**Fig. 6** The initial scene which is investigated for our exemplary scenario instantiation, and for which suitable criticality metrics are sought.

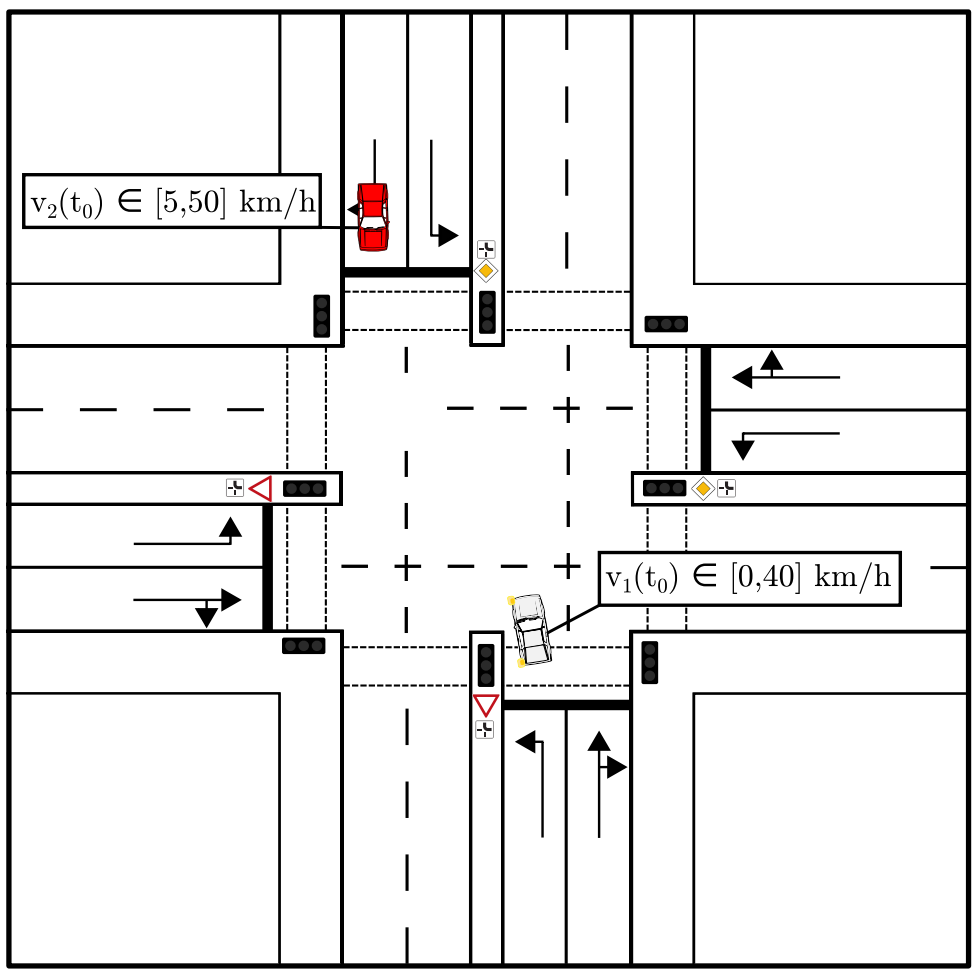

**Table 2** Derived requirements on the properties of the metrics for the exemplary application as presented in Sect. 6.2.1

| ID | Property | Description of requirement |
|---|---|---|
| $r_1$ | Subject Type | The metric shall measure the criticality from the point of view of one road vehicle to another. |
| $r_2$ | Subject Type | The metric shall be tailored towards automated vehicles, as the identified representative instances shall be critical specifically for an automation and not for human drivers. |
| $r_3$ | Scenario Type | The metric shall be applicable to urban intersections and shall be suitable for unprotected left turns. |
| $r_4$ | Inputs | The metric shall have a subset of necessary input parameters available in the simulation environment. |
| $r_5$ | Output Scale | The metric's output shall be on an ordinal scale to enable the usage of optimization algorithms. |
| $r_6$ | Reliability | The metric shall have a medium to high reliability, due to fitting a regression model on the samples. |
| $r_7$ | Validity | The metric shall have a high validity, as the critical parameter instances are used for testing and safety assurance later on. |

**Fig. 7** Visual representation of the ordering $\geq$ on the set of identified requirements $R$ within the exemplary suitability analysis

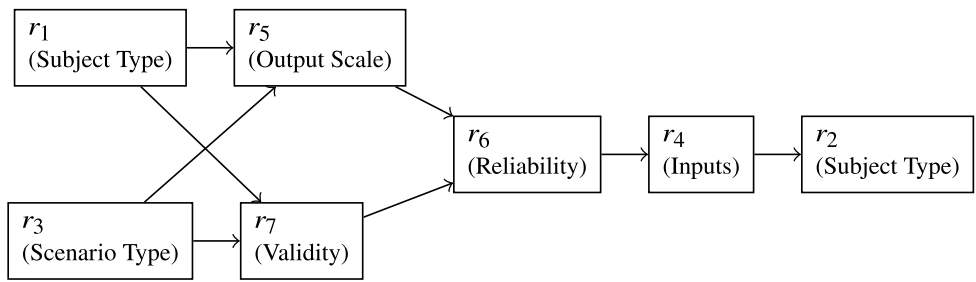

### 6.2.2 Exemplary Suitability Analysis

In the following, we walk through the analysis as presented in Sect. 6.1 using the previously introduced example application as well as the set of metrics and models $\mathcal{K}$ identified in Sect. 5 as inputs.

*Step (1)—Identification of P and R* We select the following set of properties, based on Sect. 4.1: $P$ = {Subject Type, Scenario Type, Inputs, Output Scale, Reliability, Validity}. For each $p \in P$, we identify justified requirements $r_1, \ldots, r_7 \in R$ as stated in Table 2.

*Step (2)—Ordering of R* Next, we order the requirements $r_1$ to $r_7$ according to their importance for the application. The resulting relation is depicted in Fig. 7.

Requirements $r_1$ and $r_3$ are deemed to be most important, as without suitable scenario and subject domains, the instantiation process will not yield representative results. Furthermore, $r_5$ and $r_7$ are chosen to be in the subsequent level of concern, as both the validity of the metric and its output scale are a necessity for the optimization delivering (valid) results. Finally, we order $r_6, r_4, r_2$ linearly.

*Step (3)—Determination of properties of metrics* As already indicated, for this step we use the set of metrics represented in Sect. 5.3 as a basis. This induces the set of 43 metrics, $K$ = { ACI, AGS, AM $a_{lat,req}$ $a_{long,req}$, $a_{req}$, BTN, CI, CPI, CS, DCE, $\Delta v$, DST, ET, HW, LatJ, LongJ, PET, PF, P-MC, PrET, PRI, PSD, P-SMH, P-SRS, PTTC, RSS, SOI, SP, STN, TCI, TET, THW, TIT, TTB, TTC, TTCE, TTK, TTM, TTR, TTS, TTZ, WTTC }.

According to the results of step (1), we reduce their examined properties to $P$. In favor of the conciseness of the example, we do not consider properties induced by their models, although, as shown by exemplarily examining the TTB metric, they often implicate restrictions as well. For example, certain models are unfit to assess the behavior of an AV, e.g. when including human reactions time for $A_1$, which the application defines to be automation-driven. In such a case, either the model or its parameters have to be carefully selected to validly reflect an automation's behavior, e.g. by choosing a reaction time based on the automation under test.

*Steps (4) and (5)—Iterative refinement of metrics* Firstly, Let us examine $r_1 \in \max(R)$. We now discard all metrics that are concerned with subject types other than road vehicles, namely SOI, PRI, and TTZ, as they measure criticality w.r.t. pedestrians and, at least without adaptions, are not directly suitable to assess automations.

A subsequent update of $R$ yields $\max(R) = \{r_3\}$. We therefore need to examine whether the set of remaining metrics are applicable to the intersection scenario depicted in Fig. 6. Based on the results of Sect. 5.3 and the assumption that the application prefers to use simple physics-based models, we remove any metric that

**Table 3** The resulting metrics for the exemplary suitability analysis

| Metric | Output scale |
| --- | --- |
| BTN, STN | Number |
| PET | Time |
| PrET | Time |
| $a_{req}$, $a_{long,req}$, $a_{lat,req}$ | Acceleration |
| LatJ, longj | Jerk |
| CI | Energy |
| CPI | Probability |
| P-SMH | Probability |
| P-SRS | Probability |
| P-MC | Probability |

- is primarily designed for car following or highway scenarios: ACI, DST, HW, PTTC, TCI, and THW,
- uses a predicted collision as an indicator, due to assuming a physics-based model where a collision may not be predicted over a sufficient time span in the scenario: TET, TIT, TTC, TTM, TTB, TTS, TTK, and TTR.
- is concerned with collision scenarios: $\Delta v$, or
- requires the identification of $t_{evasive}$ which may not be present in some of the scenario samples: TTA

Note that although certain metrics are removed at this point, they may be later used as a suitable enhancement factor for the resulting measurements, e.g. a multiplication with $\Delta v$.

After updating $R$, we choose $r_5 \in \max(R)$. Since we are interested in ranking scenarios w.r.t. their criticality we can not make use of nominal-scale metrics, and remove AM, RSS-DS. As a special case, we disregard SOI which reduces to a nominal scale metric for $|\mathcal{A}| = 2$.

Next, we discard $r_5$ from $R$ and select $r_7 \in \max(R)$. Considering the validity of the remainder of $K$, we find that, for criticality assessment in the exemplary scenario

- DCE does not sufficiently consider the state of the left turn vehicle, e.g. its velocity,
- ET does not model the behavior of $A_2$,
- PSD and WTTC are not able to sufficiently distinguish uncritical passing scenarios from close encounters,
- TTCE does not incorporate any semantics of the close encounter, effectively not always being a critical event,

and therefore disregard them from $K$. We also remove all metrics over which no sufficient certainty can be reached regarding their validity, namely PF and SP.

$r_6$ has no impact as no residual metric has a low reliability.

The final requirement, $r_2$, demands the metrics to be applicable to automated vehicles. It removes AGS from the resulting set, as no suitable gap acceptance models for automated vehicles exist. Furthermore, as discussed in Sect. 5.3.2, metrics such as TTB may be better suited to identify critical scenarios for humans than for AVs. The remaining metrics do not suffer from such limitations, and therefore remain in *K*. Note that this requirement can have stronger implications for selected models, which were fixed within this example. We hence complete the analysis.

*Resulting set of metrics* The iterative refinement returns the set of metrics depicted in Table 3, additionally annotated with their output scales. Depending on the planned efforts, a (potentially time-aggregated) combination of those metrics can now be chosen to fit a regression model for downstream testing purposes.

# 7 Conclusion and Future Work

We assembled an extensive, unified knowledge base on criticality metrics for automated vehicles. The totality of information has been sourced from several decades of research from the areas of traffic conflict research, traffic psychology, and, more recently, development and testing of AVs. With respect to applications of criticality metrics for automated vehicles, our method answers the question '*How to identify a set of criticality metrics suited for computing criticality within the application at hand?*'.

In order to assess the state of the art of criticality metrics, we elicited applications of criticality metrics for automated driving in Sect. 3, derived properties of such metrics in Sect. 4, analyzed the applications' requirements on them. Furthermore, we conducted an extensive review of criticality metrics in Sect. 5 and, whenever possible, separated these from the underlying models while also unifying notation. We visualized the interrelations of the considered metrics in Fig. 5. The synthesis of this review resulted in a expert-based, qualitative evaluation in Sect. 5.3 regarding the previously identified properties of metrics. On top, we sketched a methodical approach, labeled *suitability analysis*, to answer the initial research question and concluded the work at hand with an exemplary evaluation for a comprehensible yet relevant application.

Overall, this paper provides researchers and engineers working in the area of automated vehicles with a vast amount of structured and visualized information presented in a unified framework as well as a blueprint for methodically choosing suitable criticality metrics for computations within their applications.

The information gathered and the method presented in this paper offer multiple directions for future work. We finish with an outline of selected open ends.

*Quantitative evaluation of properties of criticality metrics.* For the evaluation of the properties of criticality metrics in this work, presented in Sect. 5.3, we included references whenever available. However, for many entries, we relied on expert judgment, as no evidences were traceable. In particular, quantitative studies based on a combination of synthetic and real-world data are required to confirm or refute the initial expert-based qualitative hypothesis in this work. Moreover, the influence of different models on criticality metrics and their properties has to be incorporated.

*Uncertainty quantification.* We assumed the inputs of criticality metrics to be dichotomously either available or not available. For real-world applications, however, measurements of variables are subject to measurement errors. These systematically or randomly erroneous measurements are then used to numerically approximate other variables or directly as an input to various metrics for further computations. In particular, quantification of these uncertainties is of great interest for safety-critical systems such as AVs. One option is the utilization of interval arithmetic to consistently track the numerical error coming from imprecise measurements along the metric's computations. This enables a quantification of the influence of measurement errors on a metric's output.

*Development of new metrics via formalized phenomena.* Most metrics are concerned with the physics-based symptoms of criticality – they measure spatial and temporal aspects highly associated with accidents, such as (predicted) small distances, high relative speeds, or intense decelerations. Those symptoms are typically located near the end of an accident's causal network. A more preventative approach is the identification of other causal factors, including environmental conditions, road network complexity, or traffic rule violations. Their incorporation allows to detect critical situations both earlier and more reliably. However, adequately measuring such factors necessitates their rigorous formalization. This formalization enables the considerations of more aspects of criticality, which, in conjunction, can lead to metrics that exhibit vastly improved validity and reliability.

# Appendix A: Index of Acronyms of Criticality Metrics

| | |
|---|---|
| ACI | Aggregated Crash Index |
| AGS | Accepted Gap Size |
| AM | Accident Metric |
| $a_{lat,req}$ | Required Lateral Acceleration |
| $a_{long,req}$ | Required Longitudinal Acceleration |
| $a_{req}$ | Required Acceleration |
| $a_{req,cond}$ | Conditional Required Acceleration |
| BTN | Brake Threat Number |
| CI | Conflict Index |
| CPI | Crash Potential Index |
| CrI | Crash Index |

| | | | |
|---|---|---|---|
| CS | Conflict Severity | RSS-DS | Responsibility Sensitive Safety Dangerous Situation |
| DCE | Distance of Closest Encounter | SDI | Stopping Distance Index |
| $\Delta v$ | Delta-v | SOI | Space Occupancy Index |
| DRAC | Deceleration Rate to Avoid Crash | SP | Safety Potential |
| DSSM | Deceleration-based Surrogate Safety Measure | SPrET | Scaled Predictive Encroachment Time |
| DST | Deceleration to Safety Time | STN | Steer Threat Number |
| ET | Encroachment Time | TA | Time Advantage |
| GT | Gap Time | TCI | Trajectory Criticality Index |
| HW | Headway | TET | Time Exposed TTC |
| IAPE | Initially Attempted Post Encroachment Time | THW | Time Headway |
| LatJ | Lateral Jerk | TIT | Time Integrated TTC |
| LongJ | Longitudinal Jerk | TTA | Time To Accident |
| MTTC | Modified Time To Collision | TTB | Time To Brake |
| PET | Post Encroachment Time | TTC | Time To Collision |
| PF | Potential Functions as Superposition of Scoring Functions | TTCE | Time To Closest Encounter |
| PICUD | Potential Index for Collision with Urgent Deceleration | TTK | Time To Kickdown |
| | | TTM | Time To Maneuver |
| P-MC | Collision Probability via Monte Carlo | TTO | Time To Object |
| PrET | Predictive Encroachment Time | TTR | Time To React |
| PRI | Pedestrian Risk Index | TTS | Time To Steer |
| PSD | Proportion of Stopping Distance | TTZ | Time To Zebra |
| P-SMH | Collision Probability via Scoring Multiple Hypotheses | WTTC | Worst Time To Collision |
| P-SRS | Collision Probability via Stochastic Reachable Sets | | |
| PTTC | Potential Time to Collision | | |

## Appendix B: Requirements on the Properties of Criticality Metrics

**Table 4** Minimal requirements of applications on the properties of criticality metrics

| Application | Run-time capability | Target values | Subject type | Scenario type | Inputs | Output scale | Reliability | Validity | Sensitivity | Specificity | Prediction model |
|---|---|---|---|---|---|---|---|---|---|---|---|
| A.1 Objective function | Depends on whether metrics are used in training phase or in live decision making | No, as the objective function is typically optimized | Automation, as the metric is used to evaluate potential decisions of an AV | Part of the driving function's ODD on which the metric is utilized | Output of sense component, which is the input of the planning function | Ordinal scale, to enable optimization algorithms | Medium in training phases, due to repetition of measurements, high for live decision making to enable reliable planning | High, as metric is directly or indirectly used in live decision making | High, as the AV shall not be guided into safety-critical states | High, as the AV shall adhere to performance requirements in a default driving mode | (i) Predicted maneuver duration, as safety needs to be ensured during decision horizon (ii) Branching time, as non-deterministic behavior of the AV's environment shall be considered |
| A.2 Run-tim monitoring | Yes, as the vehicle is monitored during operation | Yes, to set target values defining a safety-critical state | Automation, as the metric is used to evaluate the state an AV | Part of the driving function's ODD on which the metric is utilized | Output of sense component, which is the input of the planning function | Nominal scale, to identify safe-critical states | High, as no repeated measurements possible and safety-critical states shall be predicted reliably early on | High, as metric is directly used for live decision making | High, as monitor shall reliably detect a safety-critical state | Medium, as monitor shall not indicate unnecessarily often | (i) Minimum MRM duration, to allow some MRM to be executed if safety-critical state is predicted (ii) Branching time, as non-deterministic behavior of the AV's environment shall be considered |
| A.3 Identification of risk-reducing states | Yes, as the states need to be identified during operation | No, metric results are only compared against each other to identify an MRM | Automation, as the metric is used to evaluate potential decisions an AV | Part of the driving function's ODD on which the metric is utilized | Output of sense component, which is the input of the planning function | Ordinal scale, to compare potential MRMs against each other | High, as no repeated measurements possible and safety-critical states shall be predicted reliably early on | High, as metric is directly used in live decision making | High, to ensure correctness of MRM | Medium, as the decision space for MRMs shall not be unnecessarily reduced | (i) Predicted MRM duration, to identify a valid MRM (ii) Branching time, as non-deterministic behavior of the AV's environment shall be considered |

**Table 4** (continued)

| Application | Run-time capability | Target values | Subject type | Scenario type | Inputs | Output scale | Reliability | Validity | Sensitivity | Specificity | Prediction model |
|---|---|---|---|---|---|---|---|---|---|---|---|
| B.1.a Definin pass/fail-criteria | Depends on the test case, if no live evaluation of the criteria is planned, there exist no run-time requirements | Yes, to define the pass-/fail-conditions | Automation, specifically the vehicle under test | Test space, as criteria are employed to evaluate test cases | Depends on test bench | Nominal scale, to differentiate between pass and fail cases | High, as test cases need to be reproducible and repeated measurements are expensive | High, depending on the test context, otherwise safety goals may be violated if impact is not propagated in the safety case | High, as it is essential to not falsely label a critical test case as passed | Medium, as falsely failed test cases reduce test efficiency but not affect safety | Depends on the test case. If no live evaluation of the criteria is planned, there exist no prediction model requirements |
| B.2.a Scenario classification | No, as classification is performed offline | Yes, if classes are described using target values | Human or automation, depending on the data base or scenario space that is classified | Depends on the scenario space to classify | Depends on the purpose of the classification | Nominal scale, as classification requires only differentiation | Medium, measurements can be repeated during classification to increase reliability | Medium, depending on the purpose of classification, classes shall represent aspects of criticality to some extent | Depends on the purpose of classification | Depends on the purpose of classification | Irrelevant, as scenario classification doesnot necessitate a prediction model |
| B.2.b Scenario instantiation | No, as instantiation is performed offline | No, instantiation can be based on optimization strategies | Human or automation, depending on the scenario class to instantiate | Depends on the scenario class to instantiate | Depends on the influencing factors the instance should be representative for | Ordinal scale, to enable optimization algorithms | Medium, for low reliability, optimization algorithms may be falsely guided, but mitigation possible by increased sampling | High, due to desired representativeness of instance | Medium, to not reduce the selection space too far | High, as uncritical scenarios are undesirable instances | If class samples are already time-evaluated, no prediction is necessary. Otherwise, it needs to be evaluated in simulation, or metric shall predict with (i) Window size of typical scenario length in class (ii) Linear time |

**Table 4** (continued)

| Application | Run-time capability | Target values | Subject type | Scenario type | Inputs | Output scale | Reliability | Validity | Sensitivity | Specificity | Prediction model |
|---|---|---|---|---|---|---|---|---|---|---|---|
| B.2.c.i Selective data recording | Yes, as data needs to be selected during time of recording | Yes, due to dichotomous decision on storage of sample | Human or automation, depending on the recorder—if not an AV, human-centered metrics may be applied | Any scenario that the recording entity can possibly encounter | Positional information, as this is typically the minimal set of recorded information | Nominal scale, due to dichotomous decision on storage of sample | Low, as decreased reliability can be mitigated by implementation of hysteresis in the data recorder | Medium, validity of selected data can be increased in subsequent offline filtering or processing steps | High, as little as possible critical scenarios should be missed | Low, as it only affects efficiency of subsequent data analysis | (i) Medium, recorder needs to be triggered before critical situation has occurred (ii) Linear time, although branching time may increase validity of selected samples |
| B.2.c.ii Data filtering | No, as data is already available | Yes, due to filtering process | Human or automation, as data sets are available for both categories | Depends on the domain of the data set | Positional information over time, as this is the minimum available information in most data sets | Nominal scale, due to filtering process | Irrelevant, due to a-posteriori analysis | Medium, as not directly safety-critical | High, as little as possible critical scenarios should be missed | Low, as mostly relevant for efficiency of filtering process | Irrelevant, as traces already fully evaluated |
| B.3.a Search-based testing | No, as simulated / tested trace is evaluated a-posteriori | No, as criticality metric is optimized | Automation, specifically the vehicle under test | Test space that is aimed to be covered | Depends on the test bench | Ordinal scale, to enable optimization algorithms | High, for low reliability, optimization algorithms may be falsely guided | High, to identify critical test results | High, to lead optimization algorithm to relevant test cases | Medium, as false positives reduce test efficiency but not affect safety | No prediction model necessary, if tests are fully evaluated in simulation during testing. |
| B.3.b Test evaluation | Depends on whether tests are evaluated during time of execution | Yes, to evaluate test performance of the vehicle under test | Automation, specifically the vehicle under test | Test space that is aimed to be covered | Test run output, as the metrics are evaluated on the measured test data | Ordinal scale, due to possible ordered evaluation of test results | High, due economic factors, reduced reliability can only be mitigated by test repetition | High, depending on the test context, otherwise safety goals may be violated if impact is not propagated in the safety case | High, as it is essential to identify all critical test cases | Medium, as falsely failed test cases reduce test efficiency but not affect safety | Depends on whether tests are evaluated during time of execution |

Table 4 (continued)

| Application | Run-time capability | Target values | Subject type | Scenario type | Inputs | Output scale | Reliability | Validity | Sensitivity | Specificity | Prediction model |
|---|---|---|---|---|---|---|---|---|---|---|---|
| B.4.a Quantification for hazardous scenarios | No, as the safety case is performed offline | No, as only quantification is required | Automation, specifically the system under consideration | Any, as the quantification is performed over both the intended and actual ODD | As much as possible, due to possibly uncovered influences on scenario parameters on criticality | Ratio scale, to enable the safety case to perform comparisons between the criticality of mitigation strategies | High, due to reproducibility of safety case | High, as the quantification results are used in safety argumentation | High, as safety case argues and quantifies over *all* hazardous scenarios | Medium, as it affects efficiency but not efficacy of safety process | No prediction model necessary if argumentation is performed empirically on a data basis of scenarios |

**Supplementary Information** The online version contains supplementary material available at https://doi.org/10.1007/s11831-022-09788-7.

**Funding** Open Access funding enabled and organized by Projekt DEAL.

# References

1. Abeysirigoonawardena Y, Shkurti F, Dudek G (2019) Generating adversarial driving scenarios in high-fidelity simulators. In: 2019 international conference on robotics and automation (ICRA), pp 8271–8277. IEEE
2. Adams LD (1994) Review of the literature on obstacle avoidance maneuvers: braking versus steering. University of Michigan, Transportation Research Institute, Tech. rep
3. Alhajyaseen WK (2015) The integration of conflict probability and severity for the safety assessment of intersections. Arab J Sci Eng 40(2):421–430
4. Allen BL, Shin BT, Cooper PJ (1978) Analysis of traffic conflicts and collisions. Transp Res Rec 667:67–74
5. Althoff M, Stursberg O, Buss M (2009) Model-based probabilistic collision detection in autonomous driving. IEEE Trans Intell Transp Syst 10(2):299–310. https://doi.org/10.1109/TITS.2009.2018966
6. Archer J (2005) Indicators for traffic safety assessment and prediction and their application in micro-simulation modelling: a study of urban and suburban intersections. Ph.D. thesis, KTH Royal Institute of Technology, Stockholm, Sweden
7. Astarita V, Guido G, Vitale A, Giofré V (2012) A new micro-simulation model for the evaluation of traffic safety performances. Eur Transp Trasp Eur
8. Bagdadi O (2013) Estimation of the severity of safety critical events. Accid Anal Prev 50:167–174
9. Batsch F, Daneshkhah A, Palade V, Cheah M (2021) Scenario optimisation and sensitivity analysis for safe automated driving using Gaussian processes. Appl Sci 11(2):775
10. Bellem H, Thiel B, Schrauf M, Krems JF (2018) Comfort in automated driving: an analysis of preferences for different automated driving styles and their dependence on personality traits. Transp Res F 55:90–100. https://doi.org/10.1016/j.trf.2018.02.036
11. Boehm BW, Brown JR, Lipow M (1976) Quantitative evaluation of software quality. In: Proceedings of the 2nd international conference on software engineering, pp 592–605. IEEE
12. Broadhurst A, Baker S, Kanade T (2005) Monte Carlo road safety reasoning. In: IEEE proceedings intelligent vehicles symposium, pp 319–324. IEEE
13. Bussler A, Hartjen L, Philipp R, Schuldt F (2020) Application of evolutionary algorithms and criticality metrics for the verification and validation of automated driving systems at urban intersections. In: 2020 IEEE intelligent vehicles symposium (IV), pp 128–135. IEEE
14. Butz M, Heinzemann C, Herrmann M, Oehlerking J, Rittel M, Schalm N, Ziegenbein D (2020) SOCA: domain analysis for highly automated driving systems. In: 23rd international conference on intelligent transportation systems (ITSC), pp 1–6. IEEE (2020)
15. Cafiso S, Garcia AG, Cavarra R, Rojas MR (2011) Crosswalk safety evaluation using a pedestrian risk index as traffic conflict measure. In: Proceedings of the 3rd international conference on road safety and simulation, pp 1–15
16. Carlson WL (1979) Crash injury prediction model. Accid Anal Prev 11(2):137–153. https://doi.org/10.1016/0001-4575(79)90022-8
17. Chin HC, Quek ST (1997) Measurement of traffic conflicts. Saf Sci 26(3):169–185
18. Cunto F, Saccomanno FF (2008) Calibration and validation of simulated vehicle safety performance at signalized intersections. Accid Anal Prev 40(3):1171–1179
19. Dahl J, de Campos GR, Olsson C, Fredriksson J (2018) Collision avoidance: a literature review on threat-assessment techniques. IEEE Trans Intell Veh 4(1):101–113
20. Eggert J (2014) Predictive risk estimation for intelligent ADAS functions. In: 17th international conference on intelligent transportation systems (ITSC), pp 711–718. IEEE
21. Eidehall A (2011) Multi-target threat assessment for automotive applications. In: 14th international conference on intelligent transportation systems (ITSC), pp 433–438. IEEE
22. Evans L (1994) Driver injury and fatality risk in two-car crashes versus mass ratio inferred using newtonian mechanics. Accid Anal Prev 26(5):609–616
23. Fagnant DJ, Kockelman K (2015) Preparing a nation for autonomous vehicles: opportunities, barriers and policy recommendations. Transp Res A 77:167–181
24. Feng F, Bao S, Sayer JR, Flannagan C, Manser M, Wunderlich R (2017) Can vehicle longitudinal jerk be used to identify aggressive drivers? An examination using naturalistic driving data. Accid Anal Prev 104:125–136. https://doi.org/10.1016/j.aap.2017.04.012
25. Gabauer D, Gabler H (2006) Comparison of Delta-V and occupant impact velocity crash severity metrics using event data recorders. Annu Proc Assoc Adv Autom Med 50:57–71
26. Gangopadhyay B, Khastgir S, Dey S, Dasgupta P, Montana G, Jennings P (2019) Identification of test cases for automated driving systems using bayesian optimization. In: 22nd international conference on intelligent transportation systems (ITSC), pp 1961–1967. IEEE
27. Gladisch C, Heinz T, Heinzemann C, Oehlerking J, von Vietinghoff A, Pfitzer T (2019) Experience paper: search-based testing in automated driving control applications. In: 34th IEEE/ACM international conference on automated software engineering (ASE), pp 26–37. IEEE
28. González D, Pérez J, Milanés V, Nashashibi F (2015) A review of motion planning techniques for automated vehicles. IEEE Trans Intell Transp Syst 17(4):1135–1145
29. González L, Martí E, Calvo I, Ruiz A, Pérez J (2018) Towards risk estimation in automated vehicles using fuzzy logic. In: International conference on computer safety, reliability, and security, pp 278–289. Springer, Berlin
30. Guido G, Saccomanno F, Vitale A, Astarita V, Festa D (2011) Comparing safety performance measures obtained from video capture data. J Transp Eng 137(7):481–491. https://doi.org/10.1061/(ASCE)TE.1943-5436.0000230
31. Hallerbach S, Xia Y, Eberle U, Koester F (2018) Simulation-based identification of critical scenarios for cooperative and automated vehicles. SAE Int J Connect Autom Veh 1:93–106
32. Harman M, McMinn P (2009) A theoretical and empirical study of search-based testing: local, global, and hybrid search. IEEE Trans Softw Eng 36(2):226–247
33. Hayward JC (1972) Near miss determination through use of a scale of danger. In: 51st annual meeting of the Highway Research Board, vol 384, pp 24–34. Highway Research Board
34. Heale R, Twycross A (2015) Validity and reliability in quantitative studies. Evid Based Nurs 18(3):66–67
35. Hillenbrand J, Spieker AM, Kroschel K (2006) A multilevel collision mitigation approach-Its situation assessment, decision making, and performance tradeoffs. IEEE Trans Intell Transp Syst 7(4):528–540
36. Huber B, Herzog S, Sippl C, German R, Djanatliev A (2020) Evaluation of virtual traffic situations for testing automated driving functions based on multidimensional criticality analysis. In: 23rd international conference on intelligent transportation systems (ITSC), pp 1–7. IEEE
37. Hungar H (2020) A concept of scenario space exploration with criticality coverage guarantees. In: International symposium on leveraging applications of formal methods, pp 293–306. Springer
38. Hupfer C (1997) Deceleration to safety time (DST)–a useful figure to evaluate traffic safety? In: International cooperation of theories and concepts in Traffic Safety (ICTCT) Conference
39. Hydén C (1975) Relations between serious conflicts and traffic accidents. Tech. rep, Tekniska Högskolan i Lund, Institutionen för Trafikteknik, Lund, Sweden
40. ISO: ISO 11270:2014—Intelligent transport systems—Lane keeping assistance systems (LKAS)—performance requirements and test procedures. Standard, ISO, Geneva, Switzerland (2014)
41. ISO: ISO 15622:2018—Intelligent transport systems—adaptive cruise control systems—performance requirements and test procedures. Standard, ISO, Geneva, Switzerland (2018)
42. ISO: ISO 26262:2018: Road vehicles – Functional safety. Standard, ISO, Geneva, Switzerland (2018)
43. Jansson J (2005) Collision Avoidance Theory: With application to automotive collision mitigation. PhD Thesis, Linköping University, Linköping, Sweden
44. Jesenski S, Tiemann N, Stellet JE, Zöllner JM (2020) Scalable generation of statistical evidence for the safety of automated vehicles by the use of importance sampling. In: 23rd international conference on intelligent transportation systems (ITSC), pp 1–8 . https://doi.org/10.1109/ITSC45102.2020.9294503
45. Johnsson C, Laureshyn A, Ceunynck T (2018) In search of surrogate safety indicators for vulnerable road users: a review of surrogate safety indicators. Transp Rev 38(6):765–785
46. Joksch HC (1993) Velocity change and fatality risk in a crash-a rule of thumb. Accid Anal Prev 25:103–104
47. Junietz PM (2019) Microscopic and macroscopic risk metrics for the safety validation of automated driving. Ph.D. thesis, Technische Universität Darmstadt, Darmstadt, Germany
48. Junietz P, Bonakdar F, Klamann B (2018) PEGASUS Bericht: Kritikalitätsmetriken. Tech. rep, Institute of Automotive Engineering (FZD), Darmstadt
49. Junietz P, Bonakdar F, Klamann B, Winner H (2018) Criticality metric for the safety validation of automated driving using model predictive trajectory optimization. In: 21st international conference on intelligent transportation systems (ITSC), pp 60–65. IEEE
50. Kane A, Chowdhury O, Datta A, Koopman P (2015) A case study on runtime monitoring of an autonomous research vehicle (ARV) system. In: Runtime verification, pp 102–117. Springer
51. Klamann B, Lippert M, Amersbach C, Winner H (2019) Defining pass-/fail-criteria for particular tests of automated driving functions. In: 22nd international conference on intelligent transportation systems (ITSC), pp 169–174. IEEE

L. Westhofen et al.

52. Koopman P, Osyk B, Weast J (2019) Autonomous vehicles meet the physical world: Rss, variability, uncertainty, and proving safety. In: International conference on computer safety, reliability, and security, pp 245–253. Springer
53. Krajewski, Krajewski R, Bock J, Kloeker L, Eckstein L (2018) The highD dataset: a drone dataset of naturalistic vehicle trajectories on german highways for validation of highly automated driving systems. In: 21st international conference on intelligent transportation systems (ITSC), pp 2118–2125. IEEE
54. Kramer B, Neurohr C, Büker M, Böde E, Fränzle M, Damm W (2020) Identification and quantification of hazardous scenarios for automated driving. In: International symposium on model-based safety and assessment, pp 163–178. Springer, Berlin
55. Kruber F, Wurst J, Chakraborty S, Botsch M (2019) Highway traffic data: macroscopic, microscopic and criticality analysis for capturing relevant traffic scenarios and traffic modeling based on the highD data set. arXiv: 1903.04249
56. Kuang Y, Qu X, Wang S (2015) A tree-structured crash surrogate measure for freeways. Accid Anal Prev 77:137–148
57. Laureshyn A, Várhelyi A (2018) The Swedish traffic conflict technique: observer's manual. Lund University, Lund, Sweden, Tech. rep
58. Laureshyn A, Svensson Å, Hydén C (2010) Evaluation of traffic safety, based on micro-level behavioural data: theoretical framework and first implementation. Accid Anal Prev 42(6):1637–1646
59. Laureshyn A, Johnsson C, De Ceunynck T, Svensson Å, de Goede M, Saunier N, Włodarek P, van der Horst R, Daniels S (2016) Review of current study methods for vru safety. appendix 6 - scoping review: surrogate measures of safety in site-based road traffic observations: deliverable 2.1 - part 4. Tech. rep., InDeV, Horizon 2020 project
60. Laureshyn A, De Ceunynck T, Karlsson C, Svensson Å, Daniels S (2017) In search of the severity dimension of traffic events: extended Delta-V as a traffic conflict indicator. Accid Anal Prev 98:46–56
61. LaValle SM (2006). Planning algorithms. Cambridge University Press, Cambridge. https://doi.org/10.1017/CBO9780511546877
62. Lee YM, Madigan R, Markkula G, Pekkanen J, Merat N, Avsar H, Utesch F, Sieben A, Schießl C, Dietrich A, Boos A, Markus B, Weber F, Tango F, Portouli E (2019) interACT D.6.1. Methodologies for the evaluation and impact assessment of the interACT solutions. Dissemination report, interACT project
63. Lefèvre S, Vasquez D, Laugier C (2014) A survey on motion prediction and risk assessment for intelligent vehicles. ROBOMECH J 1(1):1–14
64. Mages M, Hopstock M, Klanner F (2009) Kreuzungsassistenz. In: Handbuch Fahrerassistenzsysteme, pp 572–581. Springer, Berlin
65. Mahmud SS, Ferreira L, Hoque MS, Tavassoli A (2017) Application of proximal surrogate indicators for safety evaluation: a review of recent developments and research needs. IATSS Res 41(4):153–163
66. Minderhoud, Michiel M, Bovy, Piet HL (2001) Extended time-to-collision measures for road traffic safety assessment. Accid Anal Prev 33:89–97
67. Morales ES, Membarth R, Gaull A, Slusallek P, Dirndorfer T, Kammenhuber A, Lauer C, Botsch M (2019) Parallel multi-hypothesis algorithm for criticality estimation in traffic and collision avoidance. In: 2019 IEEE intelligent vehicles symposium (IV), pp 2164–2171. IEEE
68. Neurohr C, Westhofen L, Henning T, de Graaff T, Möhlmann E, Böde E (2020) Fundamental considerations around scenario-based testing for automated driving. In: 2020 IEEE intelligent vehicles symposium (IV), pp 121–127. IEEE . https://doi.org/10.1109/IV47402.2020.9304823
69. Neurohr C, Westhofen L, Butz M, Bollmann MH, Eberle U, Galbas R (2021) Criticality analysis for the verification and validation of automated vehicles. IEEE Access 9:18016–18041. https://doi.org/10.1109/ACCESS.2021.3053159
70. Nistér D, Lee HL, Ng J, Wang Y (2019) The safety force field. White Paper, NVIDIA, Santa Clara, USA
71. Nonnengart A, Klusch M, Müller C (2019) CriSGen: constraint-based generation of critical scenarios for autonomous vehicles. In: International symposium on formal methods, pp 233–248. Springer, Berlin
72. Ogawa K (2007) An analysis of traffic conflict phenomenon of bicycles using space occupancy index. J Eastern Asia Soc Transp Stud 7:1820–1827
73. Oh C, Park S, Ritchie SG (2006) A method for identifying rear-end collision risks using inductive loop detectors. Accid Anal Prev 38(2):295–301
74. Ozbay K, Yang H, Bartin B, Mudigonda S (2008) Derivation and validation of new simulation-based surrogate safety measure. Transp Res Rec 2083(1):105–113
75. Peesapati LN, Hunter MP, Rodgers MO (2018) Can post encroachment time substitute intersection characteristics in crash prediction models? J Saf Res 66:205–211. https://doi.org/10.1016/j.jsr.2018.05.002
76. Perkins SR, Harris JL (1968) Traffic conflict characteristics-accident potential at intersections. Highw Res Rec pp 35–43
77. Petzoldt T (2014) On the relationship between pedestrian gap acceptance and time to arrival estimates. Accid Anal Prev 72:127–133. https://doi.org/10.1016/j.aap.2014.06.019
78. Powell J, Palacín R (2015) Passenger stability within moving railway vehicles: limits on maximum longitudinal acceleration. Urban Rail Transit 1(2):95–103. https://doi.org/10.1007/s40864-015-0012-y
79. Pütz A, Zlocki A, Bock J, Eckstein L (2017) System validation of highly automated vehicles with a database of relevant traffic scenarios. Tech. rep., 12th ITS European congress
80. Reich J, Trapp M (2020) Sinadra: towards a framework for assurable situation-aware dynamic risk assessment of autonomous vehicles. In: 2020 16th European dependable computing conference (EDCC), pp 47–50. IEEE
81. Roth M, Hendeby G, Gustafsson F (2014) Ekf/ukf maneuvering target tracking using coordinated turn models with polar/cartesian velocity. In: 17th international conference on information fusion (FUSION), pp 1–8
82. SAE: SAE J3016-201806 – Taxonomy and definitions for terms related to driving automation systems for on-road motor vehicles. Standard, SAE International, Pennsylvania (2018)
83. Schneider P, Butz M, Heinzemann C, Oehlerking J, Woehrle M (2020) Scenario-based threat metric evaluation based on the highd dataset. In: 2020 IEEE intelligent vehicles symposium (IV), pp 213–218. IEEE
84. Schneider P, Butz M, Heinzemann C, Oehlerking J, Woehrle M (2020) Scenario-based threat metric evaluation based on the highD dataset. In: 2020 IEEE intelligent vehicles symposium (IV), pp 213–218. https://doi.org/10.1109/IV47402.2020.9304726
85. Schönemann V, Winner H, Glock T, Otten S, Sax E, Boeddeker B, Verhaeg G, Tronci F, Padilla GG (2018) Scenario-based functional safety for automated driving on the example of valet parking. In: Future of information and communication conference, pp 53–64. Springer
86. Schramm D, Hiller M, Bardini R (2018) Modellbildung und Simulation der Dynamik von Kraftfahrzeugen, 3 edn. Springer, Berlin. https://doi.org/10.1007/978-3-662-54481-5
87. Schubert R, Schulze K, Wanielik G (2010) Situation assessment for automatic lane-change maneuvers. IEEE Trans Intell Transp Syst 11(3):607–616

88. Schütt B, Steimle M, Kramer B, Behnecke D, Sax E (2022) A taxonomy for quality in simulation-based development and testing of automated driving systems. In: IEEE Access, vol. 10, pp 18631-18644. https://doi.org/10.1109/ACCESS.2022.3149542
89. Shalev-Shwartz S, Shammah S, Shashua A (2017) On a formal model of safe and scalable self-driving cars. arXiv: 1708.06374
90. Sippl C, Bock F, Wittmann D, Altinger H, German R (2016) From simulation data to test cases for fully automated driving and ADAS. In: IFIP international conference on testing software and systems, pp 191–206. Springer
91. Svensson Å (1998) A method for analysing the traffic process in a safety perspective. Ph.D. thesis, Lund Institute of Technology, Lund, Sweden
92. Tak S, Kim S, Yeo H (2015) Development of a deceleration-based surrogate safety measure for rear-end collision risk. IEEE Trans Intell Transp Syst 16(5):2435–2445
93. Tam Q, Cypher-Plissart T, Ostafew CJ (2020) Proactive risk mitigation and reactive control for safe and smooth automated driving. In: RSS 2020 workshop robust autonomy
94. Tamke A, Dang T, Breuel G (2011) A flexible method for criticality assessment in driver assistance systems. In: 2011 IEEE intelligent vehicles symposium (IV), pp 697–702. IEEE . https://doi.org/10.1109/IVS.2011.5940482
95. Tsukaguchi H (1987) Mori M (1987) Occupancy indices and its application to planning of residential streets. Doboku Gakkai Ronbunshu 383:141–144
96. UL: UL 4600:2020 – Standard for Evaluation of Autonomous Products. Standard, Underwriters Laboratories, Northbrook, USA (2020)
97. Ulbrich S, Menzel T, Reschka A, Schuldt F, Maurer M (2015) Defining and substantiating the terms scene, situation, and scenario for automated driving. In: 2015 IEEE 18th international conference on intelligent transportation systems, pp 982–988. IEEE
98. Uno N, Iida Y, Itsubo S, Yasuhara S (2002) A microscopic analysis of traffic conflict caused by lane-changing vehicle at weaving section. In: Proceedings of the 13th mini-EURO conference-handling uncertainty in the analysis of traffic and transportation systems, Bari, Italy, pp 10–13
99. Van der Horst ARA (1990) A time-based analysis of road user behaviour in normal and critical encounters. Ph.D. thesis, TU Delft, Delft, Netherlands
100. Várhelyi A (1998) Drivers' speed behaviour at a zebra crossing: a case study. Accid Anal Prev 30(6):731–743
101. Wachenfeld W, Junietz P, Wenzel R, Winner H (2016) The worst-time-to-collision metric for situation identification. In: 2016 IEEE intelligent vehicles symposium (IV), pp 729–734. IEEE . https://doi.org/10.1109/IVS.2016.7535468
102. Wagner S, Groh K, Kuhbeck T, Dorfel M, Knoll A (2018) Using time-to-react based on naturalistic traffic object behavior for scenario-based risk assessment of automated driving. In: 2018 IEEE intelligent vehicles symposium (IV), pp 1521–1528. IEEE.
103. Wakabayashi H, Takahashi Y, Niimi S, Renge K (2003) Traffic conflict analysis using vehicle tracking system/digital VCR and proposal of a new conflict indicator. Infrastruct Plan Rev 20:949–956. https://doi.org/10.2208/journalip.20.949
104. Watanabe K, Kang E, Lin CW, Shiraishi S (2018) Runtime monitoring for safety of intelligent vehicles. In: Proceedings of the 55th annual design automation conference, pp 1–6
105. Wolf MT, Burdick JW (2008) Artificial potential functions for highway driving with collision avoidance. In: 2008 IEEE international conference on robotics and automation, pp 3731–3736. IEEE
106. Zofka MR, Kuhnt F, Kohlhaas R, Rist C, Schamm T, Zöllner JM (2015) Data-driven simulation and parametrization of traffic scenarios for the development of advanced driver assistance systems. In: 18th international conference on information fusion (Fusion), pp 1422–1428. IEEE